%% file: main_rev.tex
\documentclass[journal]{IEEEtran}
\ifCLASSINFOpdf
\else
\fi
\input{macros}

\usepackage{array}
\usepackage{multirow}
\usepackage{enumerate}
\usepackage{makecell}
\usepackage{tabularx} 
\usepackage{algorithm,algcompatible}
\usepackage{lipsum}
\usepackage{bbm}
\usepackage{times}
\usepackage[cmex10]{amsmath}
\usepackage{algpseudocode}
\usepackage{epsfig,verbatim}
\usepackage{amssymb} 
\usepackage{graphics,subfigure}
\usepackage{xcolor}
\usepackage{graphicx}
\usepackage{tabularx,booktabs,hhline}
\usepackage{color, colortbl}
\definecolor{LightCyan}{rgb}{0.88,1,1}
\definecolor{lightgray}{gray}{0.95}
\usepackage{multirow}

\newtheorem{theorem}{Theorem}

\newtheorem{lemma}{Lemma}

\newtheorem{remark}{Remark}

\newcommand{\argmin}{\operatornamewithlimits{argmin}}

\begin{document}
%
\title{Distributed Online Learning with Multiple Kernels}

%
%
%

\author{Jeongmin Chae,~\IEEEmembership{Student,~IEEE,}
        and~Songnam Hong,~\IEEEmembership{Member,~IEEE}
\thanks{J. Chae is with the Department of Electrical Engineering, University of Southern California, CA, 90089, USA (e-mail: chaej@usc.edu)}
\thanks{S. Hong is with the Department of Electronic Engineering, Hanyang University, Seoul, 04763, Korea (e-mail: snhong@hanyang.ac.kr)} 

}

\maketitle

\begin{abstract} 
We consider the problem of learning a nonlinear function over a network of learners in a fully decentralized fashion. Online learning is additionally assumed, where every learner receives continuous streaming data locally. This learning model is called a {\em fully distributed online learning} (or a fully decentralized online federated learning). For this model, we propose a novel learning framework with multiple kernels, which is named DOMKL. The proposed DOMKL is devised by harnessing the principles of an online alternating direction method of multipliers and a distributed Hedge algorithm. We theoretically prove that DOMKL over $T$ time slots can achieve an optimal sublinear regret $\Oc(\sqrt{T})$, implying that every learner in the network can learn a common function which has a diminishing gap from the best function in hindsight. Our analysis also reveals that DOMKL yields the same asymptotic performance of the state-of-the-art centralized approach while keeping local data at edge learners. Via numerical tests with real datasets, we demonstrate the effectiveness of the proposed DOMKL on various online regression and time-series prediction tasks.
\end{abstract}


\IEEEpeerreviewmaketitle

%
%
\section{Introduction}\label{sec:intro}


In Internet-of-Things (IoT) systems, a massive number of machine-type (or mobile) devices can be used to monitor and analyze various cyber-physical systems such as smart city, connected cars, smart factory, intelligent energy management, and so on \cite{zanella2014internet,bedi2018review}. Machine learning plays a key role in accomplishing such sophisticated tasks. In particular, this paper focuses on a nonlinear function learning as it is of great interest in various machine learning tasks such as classification, regression, clustering, and dimensionality reduction \cite{shawe2004kernel, lin2010multiple, chen2013online}. Function learning tasks, in many existing works, are conducted in a centralized fashion with the assumption that all data (possibly measured in distributed edge devices) are gathered in a central server. However, this assumption may not be acceptable due to the growing concerns of data privacy. Several cases 
of data leakage and misuse have demonstrated that centralizing local data comes at high risk for an end-user privacy. 
{ A privacy-preserving distributed learning is witnessing an unprecedented interest, in which local data is kept at distributed edge devices without centralizing the data. According to the structures of a communication network, this can be categorized into a centralized federated learning \cite{hard2018federated, sattler2019robust,  li2020federated, gursoy2016privacy} and a fully distributed learning (a.k.a., a fully decentralized federated learning) \cite{shin2017distributed, gao2015diffusion, shin2018distributed, lalitha2018fully}. In the former, a central server is used to coordinate all the participating learners during the learning process. Whereas, in the latter, edge learners can coordinate themselves to learn a function, which is the main network structure to be considered in this paper.} Furthermore, in many real-world applications, function learning tasks are expected to be performed in an online fashion. Specifically, online learning is required when data arrive in a sequential way \cite{richard2008online} and when a large number of data makes it impossible to carry out data analytic in batch form 
\cite{kivinen2004online}. Therefore, it is necessary to investigate a {\em distributed online function learning} for continuous streaming data, which is the main subject of this paper.


Learning a function involves an optimization over a function space. This challenging problem can be efficiently solved by restricting the function space to a reproducing kernel Hilbert space (RKHS) which has attractive properties from both the computational and statistical points of view \cite{scholkopf2001learning}. Such function learning approach is called a {\em kernel-based learning}. Obviously the accuracy of a kernel-based learning fully relies on a preselected basis kernel. In many real-world applications, however, it is very challenging to find an adequate single kernel. Multiple kernel learning (MKL), using a preselected set of $P$ kernels (called a kernel dictionary), is more powerful as it can enable a data-driven kernel selection from the kernel dictionary \cite{sonnenburg2006large, gonen2011multiple}. Specifically, a linear (or nonlinear) combination of multiple kernel functions is optimized as a consequence of a function learning process.

{ A kernel-based learning has been extended into a fully {\em decentralized} network \cite{shin2017distributed, gao2015diffusion, shin2018distributed} due to its necessity in various applications such as social networks, big data processing, and environmental monitoring. In \cite{shin2017distributed}, a consensus-based distributed  MKL has been proposed for regression tasks, where alternating direction method of multipliers (ADMM) \cite{boyd2011distributed}  is used as the underlying distributed optimization technique. A diffusion-based distributed learning with a single kernel has been developed in \cite{gao2015diffusion}, where the cooperation of distributed nodes is performed via the diffusion rationale. In addition, such diffusion-based approach has been incorporated into MKL framework \cite{shin2018distributed}, being able to outperform the previous works \cite{shin2017distributed, gao2015diffusion} in distributed estimation tasks. 
However, none of the above kernel-based approaches cannot operate in distribute {\em online} learning frameworks.}


In a centralized network, online kernel-based learning has been proposed, which seeks  a sequence of kernel functions from a sequential data in an online fashion \cite{kivinen2004online,sahoo2014online}. As shown in \cite{shawe2004kernel,wahba1990spline}, it suffers from a high computational complexity as the dimension of optimization variables grow with time (i.e., the number of incoming data).  This scalability problem has been addressed in \cite{shen2019random, hong2020active} via a random feature (RF) approximation \cite{rahimi2008random}. The RF-based online kernel-based learning is named online MKL (OMKL) \cite{hong2020active}, where  the optimization size can be determined irrespective of the number of incoming data. Also, as in MKL, OMKL can enjoy the advantage of using multiple kernels. Very recently, active learning strategy for OMKL has been proposed in \cite{hong2020active}, yielding an elegant accuracy performance and labeling cost tradeoff. Nevertheless, both OMKL and AMKL are not applicable to distributed online learning tasks because the underlying optimization method based on OGD  can only operate with a centralized data.  
{ As far as we know, there exists one related work for distributed kernel-based online learning, which is dubbed RFF-DOKL \cite{bouboulis2017online}. This method is devised based on single kernel-based learning with RF approximation and a diffusion strategy to satisfy the consensus constraint. As expected, it cannot yield an attractive performance mainly due to the inherent limitation of using a predetermined single kernel. Thus, it is still an open problem to construct an efficient multiple kernel-based algorithm for distributed online learning tasks, which is the primary motivation of this paper.
}

%
%
In this work, we consider a fully distributed online learning framework. In detail, we treat individual computational units (e.g., mobile devices) as learners. They aim at learning a sequence of nonlinear functions independently from their local streaming data and estimation information provided by their neighbors via a communication network. Especially, our communication network is assumed to be represented as an {\em undirected} graph, i.e., neighboring learners can communicate with each other. Inspired by the success of MKL \cite{gonen2011multiple,shin2018distributed,shen2019random,hong2020active}, we propose a novel {\em distributed online multiple kernel-based learning} (named DOMKL) for the above learning setting. { Our key contributions are summarized as follows.
\begin{itemize}
    \item In the proposed DOMKL, every learned function follows the particular structure induced by a RF-based MKL as in the centralized OMKL. Thus. it can maintain the advantages of OMKL such as scalability and an attractive accuracy. In contrast, an underlying technique to optimize the parameters of such function is completely different from that in OMKL. Thus, DOMKL can be considered as a nontrivial extension of OMKL into a fully decentralized network. 
    \item To be specific, each learned function is fully determined by the parameters of $P$ kernel functions and the weights for their proper combination. We optimize them by presenting a novel two-step approach based on online ADMM and a Hedge algorithm. Since our optimization only requires to exchange some estimates (obtained by a complex nonlinear mapping of local data) with neighboring learners, the {\em privacy} of local data is certainly preserved.
    \item We theoretically prove that DOMKL over $T$ time slots can achieve an optimal sublinear regret $\Oc(\sqrt{T})$ in terms of both learning accuracy and constraint violation (i.e., discrepancy). This implies that as time grows, every learner in a network can learn a common function having a diminishing gap from the best function in hindsight. To the best of authors' knowledge, this is the first work of its kind that proposes a multiple kernel-based algorithm with a theoretical performance guarantee under a distributed online learning setting. Specifically, our analysis reveals that DOMKL yields the same asymptotic performance with the centralized OMKL  \cite{shen2019random,hong2020active} while keeping local data at edge learners.
    \item Via numerical tests with real datasets, we demonstrate the effectiveness of the proposed DOMKL on various online regression and time-series prediction tasks. It is shown that DOMKL significantly outperforms the state-of-the-art RFF-DOKL, by enjoying the advantage of using multiple kernels. Furthermore, DOMKL can still approach the performance of the centralized OMKL, even in non-asymptotic cases. These results suggest practicality.
\end{itemize} 
}

%
%

The remaining part of this paper is organized as follows. In Section~\ref{sec:pre}, we provide some notations and definitions which will be used throughout the paper and review a multiple kernel-based online learning. The proposed DOMKL is described in Section~\ref{sec:methods}. We theoretically prove the asymptotic optimality of DOMKL in Section~\ref{sec:TA}. Beyond the asymptotic analysis, in Section~\ref{sec:Exp}, we demonstrate the effectiveness of our algorithm via experiments with real datasets.  Some concluding remarks are provided in Section~\ref{sec:con}.

{\em Notations:} Bold lowercase letters denote the column vectors. For any vector $\xv$, $\xv^{\trasp}$ denotes the transpose of $\xv$ and $\|\xv\|$ denote the $\ell_2$-norm of $\xv$. Also, $\langle\cdot,\cdot\rangle$ represents the inner product in Euclidean space. $\EE[\cdot]$ represents the expectation over an associated probability distribution. For any set $\Nc$, $|\Nc|$ denotes the cardinality of $\Nc$ (i.e., the number of elements in $\Nc$). To simplify the notations, we let $[K]\eqdef\{1,...,K\}$ for any positive integer $K$. Also, $k$, $t$, and $p$ will be used to indicate the indices of a node, time, and kernel, respectively.

%
%
\section{Preliminaries}\label{sec:pre}

In this section, we describe the basic framework of an online function learning and provide some definitions that will be used throughout the paper. The main objective of an online learning is to seek a sequence of functions $\{\hat{f}: t\in[T]\}$ such that the {\em cumulative regret} is minimized \cite{bubeck2011introduction}:
\begin{equation}\label{eq:c_regret}
    {\rm regret}(T) = \sum_{t=1}^{T} \Lc\big(\hat{f}_{t}(\xv_t), y_{t}\big) - \sum_{t=1}^{T}\Lc\big(f^{\star}(\xv_t),y_t\big),
\end{equation} where $\Lc(\cdot,\cdot)$ and $f^{\star}(\cdot)$ represent a cost (or loss) function and the best function in hindsight, respectively. Due to the nature of a streaming data, $\hat{f}_t(\cdot)$ is estimated only using the received data 
$\{(\xv_{\tau},y_{\tau}): \tau\in[t-1]\}$. Then, it will be used to generate an estimate $\hat{y}_{t} = \hat{f}_{t}(\xv_t)$ of a newly incoming data $\xv_t$. This challenging problem has been efficiently solved by incorporating a kernel-based learning into the above online learning framework \cite{kivinen2004online, shen2019random, hong2020active}. Particularly, the computational complexity of a function learning becomes tractable by restricting a function space as a well-structured reproducing Hilbert kernel space (RKHS) $\Hc_{p}$, defined as  $\Hc_{p}\eqdef\left\{f: f(\xv) = \sum_{t} \alpha_t \kappa_p(\xv,\xv_t)\right\}$, where $\kappa_p(\xv,\xv_t)$ denotes a symmetric positive semidefinite basis function (called a kernel $\kappa_p$) \cite{wainwright2019high}. One representative example is a Gaussian kernel, which is fully defined by a single parameter (called bandwidth) $\sigma_{p}^2$:
\begin{equation} \label{eq:Gaussian_kernel}
    \kappa_p(\xv,\xv_t)= \exp\left(-\frac{\|\xv-\xv_t\|^2}{2\sigma_p^2}\right).
\end{equation} 
Especially when a number of incoming data is finite,  the representer theorem in \cite{wahba1990spline} shows that an optimal solution to minimize the cumulative regret can be represented as
\begin{equation}
 \hat{f}_{t}(\xv) = \sum_{\tau=1}^{t-1} \alpha_\tau \kappa_{p}(\xv, \xv_{\tau}) \in \Hc_{p},
\end{equation} for some coefficients $\alpha_{\tau}$'s. The major drawback of the above function learning is the curse of dimensionality as the number of parameters $\alpha_{\tau}$'s (to be optimized) grows with the number of incoming data. This problem has been addressed via random feature (RF) approximation \cite{rahimi2008random} such that $\hat{f}_{t}(\xv)$ can be well-approximated as
\begin{equation}
    \hat{f}_{t}(\xv) = \hat{\thetav}_{[t,p]}^{\trasp}\zv_{p}(\xv) \in \Hc_{p},
\end{equation} with a parameter $\hat{\thetav}_{[t,p]} \in \RR^{2M\times 1}$, where the so-called randomized feature map $\zv_{p}(\cdot)$ is defined as
\begin{align}
    &\zv_p(\xv)=\nonumber\\
    &\frac{1}{\sqrt{M}}\left[\sin{\vv_1^{\trasp}\xv},...,\sin{\vv_{M}^{\trasp}(\xv),\cos{\vv_{1}^{\trasp}\xv},...,\cos{\vv_{M}^{\trasp}\xv}}\right]^{\trasp},\label{eq:zv_o}
\end{align} where $\{\vv_i: i\in[M]\}$ denotes an independent and identically distributed samples from the Fourier transform of a given kernel function $\kappa_{p}(\cdot,\cdot)$ (denoted by $\pi_{\kappa_p}(\vv)$), i.e., $\vv_i \sim \pi_{\kappa_p}(\vv)$. For a Gaussian kernel in \eqref{eq:Gaussian_kernel}, $\pi_{\kappa_p}(\vv)$ is a multivariate Gaussian distribution with the mean vector $\muv = \zerov$ and covariance matrix ${\bf \Sigma} = \sigma_{p}^{-2}\Id$. It is remarkable that via RF approximation, a function optimization has been converted into a much simper vector (or parameter) optimization. Also, the parameter $M$ is a constant not growing with a time index $t$ (e.g., $M=50$), which makes it suitable for an online learning with continuous streaming data (i.e., $t$ could be extremely large). Note that the accuracy of the kernel-based learning fully relies on a preselected kernel $\kappa_p$, which can be chosen manually either by a task-specific priori knowledge or by some intensive cross-validation process. In many real-world applications, however, finding an adequate single kernel is very demanding. Online multiple kernel learning (OMKL), using a predetermined set of $P$ kernels (called a kernel dictionary), is more powerful as it can enable a data-driven kernel selection from a given kernel dictionary \cite{shen2019random, hong2020active}. Specifically OMKL learns a sequence of functions, each of which has the form of 
\begin{equation}\label{eq:kernel_form}
    \hat{f}_t(\xv)=\sum_{p=1}^{P} \hat{q}_{[t,p]} \hat{\thetav}_{[t,p]}^{\trasp} \zv_{p}(\xv) \in \bar{\Hc},
\end{equation} where $\hat{f}_{[t,p]} (\xv)= \hat{\thetav}_{[t,p]}^{\trasp}\zv_{p}(\xv)\in \Hc_{p}$ (i.e., RKHS induced by the kernel $\kappa_p$) and $\hat{q}_{[t,p]}\in[0,1]$ denotes the combination weight of the kernel function $\hat{f}_{[t,p]}(\xv)$. { Then, we have 
\begin{align*}
    \hat{f}_t \in \bar{\Hc} &= \Hc_1 + \cdots + \Hc_{P}\nonumber\\
    &\eqdef \big\{\hat{f}_{[t,1]}+\cdots+\hat{f}_{[t,P]}: \hat{f}_{[t,p]} \in \Hc_{p}\big\},
\end{align*} and $\bar{\Hc}$ is an RKHS from \cite[Proposition 12.27]{wainwright2019high}.}

Motivated by the success of multiple kernel-based learnings \cite{shin2017distributed, shin2018distributed,shen2019random, hong2020active}, our learned function in Section~\ref{sec:methods} will be assumed as a parameterized function in (\ref{eq:kernel_form}).  { We emphasize that the parameters to be optimized in the proposed algorithms are equivalent to those in OMKL \cite{hong2020active} (e.g., $\{\hat{\thetav}_{[t,p]}, \hat{q}_{[t,p]}: p \in [P]\}$), whereas the optimization technique (or learning algorithm) in Section~\ref{sec:methods} is completely different from that in OMKL.}


\section{Methods}\label{sec:methods}

In this section, we study the problem of an online learning over a {\em fully decentralized network} consisting of a set of $K$ learners (or nodes), indexed by $k \in \Vc=\{1,2,...,K\}$. They learn a sequence of nonlinear functions independently from their local streaming data and pass estimated information to their neighbors in a communication network. { This network is completely defined by an {\em undirected} connected graph $\Gc=(\Vc,\Ec)$. For the case of a disconnected graph, the proposed algorithms can be applied to each connected component separately.} The $\Ec$ represents the set of {\em unordered} pairs of the learners (called edges), defined as $\Ec=\{\{k,\ell\}: \mbox{the nodes } k \mbox{ and }  \ell \mbox{ are connected}\}$.
This shows the connectivity of the $K$ learners (i.e., a network structure). For any node $k \in \Vc$, the index subset of its neighbors is defined as $\Nc_{k} = \{\ell: \{k,\ell\} \in \Ec\}\subseteq \Vc$.
Since an undirected graph is assumed, it is obvious that $\ell \in \Nc_{k}$ and $k \in \Nc_{\ell}$ if $\{k,\ell\}\in\Ec$. Given a communication network $\Gc=(\Vc,\Ec)$, it is assumed that at every time $t$, each learner $k \in \Vc$ can transmit the latest estimated information $\{\hat{\thetav}_{[k,t,p]}, \hat{q}_{[k,t,p]} :p\in[P]\}$ to its neighbors $\ell \in \Nc_k$. Definitely this communication protocol can preserve privacy of local data as it should not be recovered from the shared parameters and the relationship in \eqref{eq:kernel_form} (i.e., the non-linearity of our function model).

Given a communication network $\Gc=(\Vc,\Ec)$, our goal is to optimize the parameters of $P$ kernel functions and the combination weights in a distributed way. In Section~\ref{subsec:single}, we first introduce a distributed online learning with a predetermined single kernel (named DOKL). Appropriately combining this method with a distributed Hedge algorithm, in Section~\ref{subsec:multiple},  we propose a novel distributed online learning with multiple kernels (named DOMKL).



%
%
\subsection{The Proposed DOKL}\label{subsec:single}

We first introduce a distributed kernel learning and extend it into our {\em online} learning framework. { Without loss of generality, a single kernel $\Hc_{p}$, defined by a kernel $\kappa_{p}$, is assumed. As shown in \eqref{eq:zv_o}, due to the use of RF approximation, the characteristic of the kernel $\kappa_p$ is only reflected by a randomized feature map $\zv_{p}(\cdot)$. Using the same seed value to generate the $\zv_{p}(\cdot)$, all distributed learners can have the identical map $\zv_{p}(\cdot)$. Recall that $M$ and $K$ indicate the size of random features (see \eqref{eq:zv_o}) and the size of the set $\Vc$ (i.e., $|\Vc|=K)$, respectively.} In a distributed kernel learning, every learner $k \in \Vc$ has its own local dataset $\{(\xv_{k,1},y_{k,1}),...,(\xv_{k,T},y_{k,T})\}$ and learns a function $\hat{f}_{k}(\xv)$ collaboratively with its neighbors subject to a consensus constraint. Also, based on a kernel-based learning, each function $\hat{f}_{k}$(\xv) is fully determined by a parameter $\hat{\thetav}_{[k,p]} \in \RR^{2M\times 1}$, i.e., 
\begin{equation}\label{eq:single_k}
    \hat{f}_{k}(\xv) \eqdef \hat{\thetav}_{[k,p]}^{\trasp} \zv_{p}(\xv),\; \forall k \in \Vc.
\end{equation} The parameters $\{\hat{\thetav}_{[k,p]}: k\in \Vc\}$ in \eqref{eq:single_k} can be optimized by taking the solution of 
\begin{align}
   &\argmin_{\{\thetav_k:k\in\Vc\}}\;\; \sum_{t=1}^T \sum_{k=1}^{K} \Lc(\thetav_{k}^{\trasp}\zv_p(\xv_{k,t}), y_{k,t})\nonumber\\
    &\mbox{ subject to }\;\; \thetav_{k} = \thetav_{\ell}, \forall k\in\Vc, \ell\in \Nc_k. \label{eq:const_o1}
\end{align}  The above constraint is known as {\em consensus constraint} in the context of a distributed optimization.
In \cite{boyd2011distributed}, it was shown that the above problem can be solved in a distributed way via alternating direction method of multipliers (ADMM). Leveraging this, we convert the optimization problem \eqref{eq:const_o1} into the standard from of ADMM:
\begin{align}
     &\argmin_{\{\thetav_k:k\in\Vc\}}\;\; \sum_{t=1}^T \sum_{k=1}^{K} \Lc(\thetav_{k}^{\trasp}\zv_p(\xv_{k,t}), y_{k,t})\nonumber\\
    &\mbox{ subject to }\;\; \thetav_{k} = \gammav_{\{k,\ell\}},\;\; \forall k \in \Vc, \ell \in \Nc_{k},  \label{eq:const_o2}
\end{align} where $\gammav_{\{\ell,k\}}=\gammav_{\{\ell,k\}}$ denotes an auxiliary vector assigned to every edge $\{k,\ell\}\in\Ec$. It is easily verified that the constraints in (\ref{eq:const_o2}) and (\ref{eq:const_o1}) are equivalent. 

We are now ready to propose a kernel-based distributed online learning (named DOKL). The proposed DOKL aims at solving the optimization problem (\ref{eq:const_o2}) in an {\em online} fashion, under the assumption that data arrives sequentially. From our theoretical analysis in Section~\ref{sec:TA}, it is proved that the optimal sublinear regrets from the optimal performance of \eqref{eq:const_o2} can be achieved, if the following optimization problem is solved exactly at every time $t$:
\begin{align}
    &\{\hat{\thetav}_{[t+1,k,p]}:k\in\Vc\} =\nonumber\\
    &\;\; \argmin_{\{\thetav_k:k\in\Vc\}}\;\;  \sum_{k=1}^{K} \Lc(\thetav_k^{\trasp}\zv_p(\xv_{k,t}), y_{k,t}) + \frac{\eta_l}{2}\sum_{k=1}^{K}\|\thetav_k - \hat{\thetav}_{[k,t,p]}\|^2\nonumber\\
    &\;\;\mbox{subject to }\;\; \thetav_{k} = \gammav_{\{k,\ell\}},\;  \in \forall k \in \Vc, \ell \in \Nc_k, \label{eq:opt1}
\end{align} where $\eta_l > 0 $ denotes a learning rate.
Moreover, Lemma~\ref{lem1-OADMM} shows that the above problem can be solved exactly in a distributed way, by harnessing the principle of {\em online} ADMM \cite{wang2013online}. Therefore, the proposed DOKL based on Lemma~\ref{lem1-OADMM} can solve the optimization problem  (\ref{eq:const_o2}) in a distributed way while ensuring an asymptotic optimality. 
%
%
\vspace{0.1cm}
\begin{lemma}\label{lem1-OADMM} The optimization problem in \eqref{eq:opt1} can be solved in a distributed way, wherein each learner $k$ only exchanges the 
$\hat{\thetav}_{[k,t,p]}$ and $\hat{\thetav}_{[\ell,t,p]}$
with its neighbors $\ell \in \Nc_k$. Given parameters $\eta_l>0$ and $\rho>0$, the corresponding updates at the learner $k$ are given as
\begin{align}\hspace{-0.1cm}
    &\hat{\thetav}_{[k,t+1,p]}=\argmin_{\thetav_k} \Lc(\thetav_{k}^{\trasp}\zv_p(\xv_{k,t}), y_{k,t}) + \hat{\lambdav}_{[k,t,p]}^{\transp}\thetav_k \nonumber\\
    &+\frac{\rho}{2}\sum_{\ell \in \Nc_k}\left\|\thetav_k - \frac{\hat{\thetav}_{[k,t,p]}+\hat{\thetav}_{[\ell,t,p]}}{2}\right\|^2 + \frac{\eta_{l}}{2}\|\thetav_k - \hat{\thetav}_{[k,t,p]}\|^2,\label{eq:theta-update}\\
  &\hat{\lambdav}_{[k,t+1,p]} =\hat{\lambdav}_{[k,t,p]} + \frac{\rho}{2}\sum_{\ell \in\Nc_k}(\hat{\thetav}_{[k,t+1,p]}-\hat{\thetav}_{[\ell,t+1,p]}). \label{eq:lambda-update}
\end{align} 
\end{lemma}
\begin{IEEEproof} The proof is provided in Appendix~\ref{app:lem1}.
\end{IEEEproof} { Note that a regularization parameter $\rho$ can control the tradeoff between the loss and discrepancy (i.e., constraint violation). The rationale to choose the hyper-parameters $\eta_l$ and $\rho$ of DOKL will be provided in 
Section~\ref{sec:TA} via theoretical analysis.} From Lemma~\ref{lem1-OADMM}, at time $t$, every learner $k \in \Vc$ performs the following procedures in parallel:
\begin{itemize}
    \item The learner $k$ stores the latest estimates  $\hat{\thetav}_{[k,t,p]}$ and $\hat{\lambdav}_{[k,t,p]}$ at the previous time.
    \item Using them, it locally updates the parameter $\hat{\thetav}_{[k,t+1,p]}$ by solving the  optimization problem (\ref{eq:theta-update}). See Remark~\ref{rem:opt} for the case of quadratic loss function.
    \item Then, it transmits the $\hat{\thetav}_{[k,t+1,p]}$ to and receives $\{\hat{\thetav}_{[\ell,t+1,p]}: \ell \in \Nc_{k}\}$ from the neighbors.
    \item Leveraging updated estimates, the learner $k$ updates the discrepancy $\hat{\lambdav}_{[k,t+1,p]}$ locally.
\end{itemize} The above procedures are summarized in Algorithm 1.


\vspace{0.1cm}
\begin{remark}\label{rem:opt} Suppose that the {\em quadratic} loss function is used:
\begin{align}
    \Lc(\thetav^{\transp}\zv(\xv_t),y_t) = (\thetav^{\transp}\zv(\xv_{t})-y_{t})^2.
\end{align} In this case, the closed-form solution of (\ref{eq:theta-update}) is obtained as
\begin{align*}
    \hat{\thetav}_{[k,t+1,p]} &= \Big(2\zv(\xv_{k,t})\zv(\xv_{k,t})^{\transp} + (\eta_l+\rho|\Nc_k|)\Id\Big)^{-1}\nonumber\\
    & \times \left(2y_{k,t}\zv(\xv_{k,t}) +\eta_{l}\hat{\thetav}_{[k,t,p]} + \rho\hat{\gammav}_{[k,t,p]} - \hat{\lambdav}_{[k,t,p]}\right),
\end{align*} where $\hat{\gammav}_{[k,t,p]}=\sum_{\ell \in \Nc_k} (\hat{\thetav}_{[k,t,p]}+\hat{\thetav}_{[\ell,t,p]})/2$. This closed-form expression will be used for our experiments in Section~\ref{sec:Exp}.
\end{remark}

\begin{algorithm}
\caption{DOKL (at the learner $k \in \Vc$)}
\begin{algorithmic}[1]

\State {\bf Input:} Network $\Gc=(\Vc,\Ec)$, a preselected kernel $\kappa_p$, and hyper-parameters $(\rho, \eta_l, M)$. 
\State {\bf Output:} A sequence of functions $\hat{f}_{k,t}(\xv),\; t\in [T+1]$.
\vspace{0.05cm}

\State {\bf Initialization:} $\hat{\thetav}_{[k,1,p]} = \zerov$ and $\hat{\lambdav}_{[k,1,p]} = \zerov$. Each learner constructs the identical random feature map $\zv_{p}(\cdot)$ with the same seed.

\vspace{0.05cm}
\State {\bf Iteration:} $t=1,...,T$
\begin{itemize}
    \item Receive a streaming data $(\xv_{k,t},y_{k,t})$.
    \item Construct $\zv_p(\xv_{k,t})$ via (\ref{eq:zv_o}) using the kernel $\kappa_p$.
    \item Update $\hat{\thetav}_{[k,t+1,p]}$ from (\ref{eq:theta-update}). 
    \item Transmit $\hat{\thetav}_{[k,t+1,p]}$ to and receive $\hat{\thetav}_{[\ell,t+1,p]}$ from the neighbors $\ell \in \Nc_k$.
    \item Update $\hat{\lambdav}_{[k,t+1,p]}$ via (\ref{eq:lambda-update}).
\end{itemize}
\end{algorithmic}
\end{algorithm}

%
%
\subsection{The Proposed DOMKL}\label{subsec:multiple}

The proposed DOKL in Section~\ref{subsec:single} has an inevitable limitation of using a predetermined single kernel. This problem can be even worse in an online learning framework as the single kernel should be determined without observing a sequential data. Thus, it is required to extend DOKL into a multiple kernel setting. Motivated by this, we propose a distributed online multiple kernel-based online learning (DOMKL). Hereinafter, it is assumed that there are $P$ kernels in a kernel dictionary. Under RF approximation, each kernel $p$ is specified by a kernel-dependent mapping $\zv_{p}(\xv)$ in \eqref{eq:zv}, for $p \in [P]$.
{ It is remarkable that we will not formulate an optimization for DOMKL (i.e., online ADMM formulation for the parameters of a multiple kernel-based function in \eqref{eq:DOMKL_f}), since the resulting optimization seems to be quite complex. Instead, we resort to a simpler two-step approach, wherein the parameters of $P$ local functions and their combination weights are optimized separately. It will be proved in Section~\ref{sec:TA} that our approach can guarantee an order-optimal performance.} Then, the proposed DOMKL is performed with the following two steps.

\vspace{0.1cm}
\noindent{\em i) Local step:} In this step, each learner $k$ optimizes the parameters of $P$ kernel functions (denoted by $\{\hat{\thetav}_{[k,t+1,p]}: p\in[P]\}$) independently from each other. Given a kernel $\kappa_p$, such optimization has been already solved in Section~\ref{subsec:single}, namely, $\hat{\thetav}_{[k,t+1,p]}$ is obtained via \eqref{eq:theta-update}. From this, every kernel $k$ can simply obtain the updated parameters of the $P$ kernel functions.

\vspace{0.1cm}
\noindent{\em ii) Global step:} In this step, each learner $k$ seeks the best function approximation $\hat{f}_{[k,t+1]}(\xv)$ by combining its kernel functions $\{\hat{f}_{[k,t+1,p]}(\xv), p\in [P]\}$ with proper weights $\{\hat{q}_{[k,t+1,p]}, p\in[P]\}$:
\begin{equation}\label{eq:DOMKL_f}
    \hat{f}_{[k,t+1]}(\xv) = \sum_{p=1}^{P} \hat{q}_{[k,t+1,p]}\hat{f}_{[k,t+1,p]}(\xv),
\end{equation} where $\sum_{p=1}^P \hat{q}_{[k,t+1,p]} = 1$ with $\hat{q}_{[k,t+1,p]} \in [0,1]$. Thus, we need to optimize the combination weights so that they can capture the reliabilities of the $P$ kernels. Based on the principle of exponential strategy (or Hedge algorithm) \cite{bubeck2011introduction}, each learner $k$ determines the weights $\{\hat{q}_{[k,t+1,p]}:p\in[P]\}$ on the basis of its own and neighbors' past losses. Towards this, we first define the accumulated losses of the learner $k$: for $p \in [P]$, 
\begin{equation}
    \hat{w}_{[k,t+1,p]} = \exp\left(-\frac{1}{\eta_g} \sum_{\tau=1}^t \Lc(\hat{f}_{[k,\tau,p]}(\xv_{k,\tau}), y_{k,\tau}) \right), \label{eq:local-w}
\end{equation} with some learning rate $\eta_g > 0$. Combining with $\{\hat{w}_{[\ell,t+1,p]}: \ell \in \Nc_k\}$ (provided by the neighbors), the weights are computed as
\begin{equation}
    \hat{q}_{[k,t+1,p]} = \frac{\hat{w}_{[k,t+1,p]}\times \prod_{\ell\in \Nc_k}\hat{w}_{[\ell,t+1,p]}}{\sum_{p=1}^P \hat{w}_{[k,t+1,p]}\times \prod_{\ell \in\Nc_k}\hat{w}_{[\ell,t+1,p]}}.\label{eq:weight-update}
\end{equation}The learned function in (\ref{eq:DOMKL_f}) with the above optimized parameters will be used to estimate the label $\hat{y}_{k,t+1}$ of a newly incoming data $\xv_{k,t+1}$, i.e., 
\begin{equation}
    \hat{y}_{k,t+1} = \hat{f}_{[k,t+1]}(\xv_{k,t+1}).
\end{equation} The procedures of DOMKL are summarized in Algorithm 2.

\vspace{0.1cm}
\begin{remark} We will verify that the proposed weight-update in \eqref{eq:weight-update} is performed in a network-wise, thus enabling to satisfy the consensus constraint. From the update rule in \eqref{eq:weight-update}, we can have:
\begin{align*}
    \hat{w}_{[k,t+1,p]}\times \prod_{\ell \in \Nc_k}\hat{w}_{[\ell,t+1,p]} = \exp\left(-\frac{1}{\eta_g}L_{[k,t+1,p]}\right),
\end{align*} where the accumulated losses $L_{[k,t+1,p]}$ can be expressed as
\begin{align}
    L_{[k,t+1,p]} &= \sum_{\tau=1}^{t} \Lc(\hat{f}_{[k,\tau,p]}(\xv_{k,\tau}), y_{k,\tau}) \nonumber\\
    &\;\;\;\;\;\;\;\;\;\;\;\;+\sum_{\ell \in \Nc_k} \sum_{\tau=1}^{t} \Lc(\hat{f}_{[\ell,\tau,p]}(\xv_{\ell,\tau}), y_{\ell,\tau}).\label{eq:second}
\end{align} The second term in \eqref{eq:second} captures the reliability of the kernel $p$ on the basis of the local data of neighboring learners.
\end{remark}
\vspace{0.1cm}
\begin{remark}
When a network graph $\Gc$ forms an {\em acyclic graph}, the combination weights can be further elaborated by exploiting the principle of message-passing \cite{mezard2009information}. Specifically, for acyclic networks, the weights can be updated using the local data of all connected learners. Before stating the update rule, some useful definitions are first provided. We say that two learners are connected with a length $\zeta$ if there exists an undirected path of length $\zeta$ between these nodes. Also, let $\Nc_{k}^{(\zeta)}$ be the index set containing the length-$\zeta$ connected learners of the learner $k$, i.e., $\Nc_{k}^{(\zeta)}=\{\ell \in\Vc: \mbox{there exists a length-$\zeta$ path between the learners $k$ and $\ell$}\}$, where $\Nc_{k}^{(1)} = \Nc_k$. Harnessing the principle of message-passing, the message transmitted from the learner $k$ to its neighbors $\ell \in \Nc_k$ is determined as
\begin{equation}
    m_{k \rightarrow \ell,t+1} = \hat{w}_{[k,t+1,p]}\times \prod_{i \in \Nc_k: i \neq \ell} m_{i \rightarrow k, t}
\end{equation} with initial values $m_{i \rightarrow k,1}=1$ for all $\{i,k\}\in \Ec$. Accordingly, the weights in the learner $k$ are updated using the incoming messages $\{m_{\ell\rightarrow k, t}: \ell \in \Nc_{k}\}$ such as
\begin{equation}
    \hat{q}_{[j,t+1,p]} = \frac{\hat{w}_{[j,t+1,p]}\times \prod_{\ell \in \Nc_k} m_{\ell\rightarrow k, t}}{\sum_{p=1}^P \hat{w}_{[k,t+1,p]}\times \prod_{\ell \in \Nc_k} m_{\ell\rightarrow k, t}  },
\end{equation} for some parameter $\eta_g>0$. This update rule can guarantee that the information of local losses can be propagated over the entire network as a time (or iteration) $t$ grows, which is manifested as follows:
\begin{equation}
    \hat{w}_{[k,t+1,p]}\times \prod_{\ell \in \Nc_k} m_{\ell\rightarrow k, t} = \exp\left(-\frac{1}{\eta_g}L_{[k, t+1,p]}\right),
\end{equation} where the accumulated losses are computed as
\begin{align}
     L_{[k, t+1,p]} &= \sum_{\tau=1}^{t} \Lc(\hat{f}_{[k,\tau,p]}(\xv_{k,\tau}), y_{k,\tau})\nonumber\\
     &\;\;\;+\sum_{\zeta=1}^{t-1}\sum_{\ell \in \Nc_k^{(\zeta)}} \sum_{\tau=1}^{t-\zeta} \Lc(\hat{f}_{[\ell,\tau,p]}(\xv_{\ell,\tau}), y_{\ell,\tau}).
\end{align} This message-passing update can be used even for cyclic graphs, provided that the girth of a network graph $\Gc$ is larger than the number of incoming data $T$. Otherwise, some local losses can be reflected because of the duplication.
\end{remark}

{
\begin{remark} Definitely, our approach in Section~\ref{subsec:multiple}  can be naturally applied to another single kernel-based algorithm (named RFF-DOKL)
proposed in \cite{bouboulis2017online}. Then, the difference is in the underlying distributed optimization methods. Specifically, online ADMM is used in the proposed DOKL while RFF-DOKL is based on online gradient descent (OGD and a diffusion strategy. Although such extension would be straightforward with respect to algorithm, it requires an more effort to prove the asymptotic optimality of the resulting method. The proof is quite demanding and left for an interesting future work.
\end{remark}
}

\begin{algorithm}
\caption{DOMKL (at the learner $k \in \Vc$)}
\begin{algorithmic}[1]
\State {\bf Input:} Network $\Gc=(\Vc,\Ec)$, a preselected set of $P$ kernels $\kappa_p$, $p\in[P]$ and hyper-parameters $(\rho, \eta_l, \eta_g, M)$.  
\State {\bf Output:} A sequence of functions $\hat{f}_{[k,t]}(\xv)$, $t\in [T+1]$.
\State {\bf Initialization:}  $\hat{\thetav}_{[k,1,p]} = \zerov$, $\hat{\lambdav}_{[k,1,p]}=\zerov$ and $\hat{w}_{[k,1,p]} = 1, \forall p\in[P]$. Each learner constructs the identical random feature map $\zv_{p}(\cdot)$, $\forall p\in [P]$ with the same seed.
\State {\bf Iteration:} $t=1,...,T$
\begin{itemize}
    \item Receive a streaming data $(\xv_{k,t},y_{k,t})$.
    \item Construct $\zv_{p}(\xv_{k,t})$ via (\ref{eq:zv_o}) for $\kappa_p, \forall p\in[P]$.
    \item Update $\hat{\thetav}_{[k,t+1,p]}$ via (\ref{eq:theta-update}) for $\forall p\in[P]$.
    \item Update $\hat{w}_{[k,t+1,p]}$ via (\ref{eq:local-w}) for $\forall p\in[P]$.
    \item Transmit $\{\hat{\thetav}_{[k,t+1,p]}, \hat{w}_{[k,t+1,p]}\}$ to and receive $\{\hat{\thetav}_{[\ell,t+1,p]},\hat{w}_{[\ell,t+1,p]}\}$ from the neighbors $\forall \ell \in \Nc_k$, for $\forall p\in[P]$.
    \item Update $\hat{q}_{[k,t+1,p]}$ via (\ref{eq:weight-update}).
    \item Update $\hat{\lambdav}_{[j,t+1,p]}$ via (\ref{eq:lambda-update}).
    \item Update $\hat{f}_{[k,t+1]}(\xv) = \sum_{p=1}^{P} \hat{q}_{[k,t+1,p]}\hat{\thetav}_{[k,t+1,p]}^{\transp}\zv_{p}(\xv)$.
\end{itemize}
\end{algorithmic}
\end{algorithm}

%
%
\section{Regret Analysis}\label{sec:TA}

In this section, we analyze the performances of the proposed DOKL and DOMKL in terms of learning accuracy and constraint violation (i.e., discrepancy). These cumulative regrets are commonly used for the analysis of online distributed optimizations \cite{wang2013online}. Let $\hat{f}_{[k,t]}$ denote the estimated function of the learner $k$. Also, let $f_{k}^{\star}$ denote the optimal function in terms of the incoming data at the learner $k$, namely,
\begin{equation}
    f_{k}^{\star} = \argmin_{f \in \bar{\Hc}} \sum_{t=1}^{T}\Lc(f(\xv_{k,t}),y_{k,t}),
\end{equation}  
subject to the consensus constraint $f_k^{\star} = f_\ell^{\star}$ for $\forall k \in \Vc, \ell \in \Nc_k$. For the assumption of a connected graph $\Gc=(\Vc,\Ec)$, the consensus constraint implies that $f_{k}^{\star}=f^{\star}$ for all $k \in \Vc$.
As in \cite{wang2013online}, the cumulative regrets for learning accuracy and consensus violation (or discrepancy) at the node $k$ are respectively defined as
\begin{align*}
    {\rm regret}_{\rm a}^k(T) &=\sum_{t=1}^{T} \Lc(\hat{f}_{[k,t]}(\xv_{k,t}),y_{k,t})-\Lc\left(f_{k}^{\star}(\xv_{k,t}),y_{k,t}\right)\\
    {\rm regret}_{\rm d}^k (T) &=\sum_{t=1}^{T} \left[\sum_{\ell\in \Nc_k}\hat{f}_{[k,t]}(\xv_{k,t}) -\hat{f}_{[\ell,t]}(\xv_{k,t})\right]^2.
\end{align*}

From now on, we will prove that the proposed DOKL and DOMKL can achieve the optimal sublinear regrets $\Oc(\sqrt{T})$ for both learning accuracy and discrepancy. Namely, as $T \rightarrow \infty$, DOKL and DOMKL can achieve negligible gaps from the associated optimal performances. We remark that in general, the optimal performance of DOMKL is likely to be much better than that of DOKL, because of the advantage of using multiple kernels. Before stating our main results, some useful notations and definitions will be provided. Recall that in the proposed methods in Section~\ref{sec:methods}, each kernel function $\hat{f}_{[k,t,p]}$ is represented as
\begin{equation}
    \hat{f}_{[k,t,p]}(\xv) = \hat{\thetav}_{[k,t,p]}^{\trasp}\zv_{p}(\xv).
\end{equation} 
For ease of exposition, given the data $(\xv_{k,t}, y_{k,t})$, the loss function with respect to $\thetav$ is defined as 
\begin{equation}
   \Lc_{[k,t,p]}(\thetav) = \Lc(\thetav^{\trasp}\zv_p(\xv_{k,t}),y_{k,t}).
\end{equation} Hereinafter, the above two notations will be used interchangeably. 
Let $\thetav_{[k,p]}^{\star}$ be the parameter of an optimal kernel function in the kernel $\kappa_p$, i.e.,
\begin{equation}
    \thetav_{[k,p]}^{\star} = \argmin_{\thetav} \sum_{t=1}^{T}\Lc_{[k,t,p]}(\thetav), \forall k \in \Vc.
\end{equation} subject to the consensus constraint in (\ref{eq:opt1}). As noticed before, when $\Gc=(\Vc,\Ec)$ is a connected graph, we have that $\thetav_{p}^{\star}=\thetav^{\star}_{[k,p]}$ for all $k \in \Vc$. For our analysis, the following conditions are assumed:
\begin{itemize}
    \item {\bf (a1)} For any fixed $\zv_{p}(\xv_{k,t})$ and $y_{k,t}$, the loss function $\Lc_{[k,t,p]}(\thetav)$ is convex with respect to $\thetav$, differentiable, and bounded as  $\Lc_{[k,t,p]}(\thetav) \in [0,L_u]$. Also, its gradient is bounded, i.e., $\|\nabla \Lc_{[k,t,p]}(\thetav)\|^2 \leq G$.\vspace{0.2cm}
    \item {\bf (a2)} For any kernel $\kappa_p$, $\hat{\thetav}_{[k,t,p]}$ belongs to a bounded set $\Theta_p \subseteq \RR^{2D\times 1}$, i.e., $\|\hat{\thetav}_{[k,t,p]}\|^2\leq C$.\vspace{0.2cm}
    \item {\bf (a3)} For any time $t$, $\Lc_{[k,t,p]}(\hat{\thetav}_{[k,t,p]}) - \Lc_{[k,t,p]}(\thetav_{[k,p]}^{\star}) \geq  -  B$
    for some positive constant $B$.\vspace{0.2cm}
     \item {\bf (a4)} For any fixed $k \in \Vc$ and $p \in [P]$, there exists a sequence of $\epsilon_t$'s such that $|\hat{q}_{[k,t,p]} - \hat{q}_{[\ell,t,p]}|\leq \epsilon_t$ for all $\ell \in \Nc_k$ and $\sum_{t=1}^T \epsilon_t^2 \leq \Oc(\sqrt{T})$.
\end{itemize} The assumptions {\bf (a1)} and {\bf (a2)} are generally required for the analysis of online learning setting \cite{shen2019random,hong2020active, hazan2016introduction, wang2013online}. Also, the assumption {\bf (a3)} is required to prove the sublinear regret of constraint violation (i.e., discrepancy), which is true if convex functions are bounded from below or Lipschitz continuous \cite{wang2013online}. The assumption {\bf (a4)} is required for the proof of Theorem~\ref{thm2}, which implies that the degree of heterogeneity of distributed local data is bounded and thus results in a bounded loss (e.g., $\Oc(\sqrt{T})$ of the proposed DOMKL. 
In fact, the upper-bound $\epsilon_t$ is determined on the basis of network structure (i.e., the connectivity of nodes in the network). For example, when the network is a complete graph, we can easily obtain the $\epsilon_t = 0$ for all $t \in [T]$. Obviously, as the connectivity of a network becomes sparse, $\epsilon_t$ tends to increase. Via numerical tests, we have confirmed that the assumption {\bf (a4)} might not be tight, i.e., it can be easily satisfied in practical network structures.

We first state the main results of this section in Theorems~\ref{thm1} and~\ref{thm2} below, and the proofs will be provided in Sections~\ref{subsec:proof1} and~\ref{subsec:proof2}.
%
%
\vspace{0.1cm}
\begin{theorem}\label{thm1} Under the assumptions {\bf (a1)} - {\bf (a3)}, DOKL in Algorithm 1 with the parameters $\rho=\eta_l=\Oc(\sqrt{T})$ and any preselected kernel $\kappa_p$ can achieve the sublinear regrets  as
\begin{align*}
    &{\rm regret}_{\rm a}^k(T) \\
    &\;=\sum_{t=1}^{T} \Lc_{[k,t,p]}(\hat{\thetav}_{[k,t,p]})-\sum_{t=1}^{T}\Lc_{[k,t,p]}(\thetav_{[k,p]}^{\star})\leq \Oc(\sqrt{T}),\\
    &{\rm regret}_{\rm d}^{k}(T)\\
    &\; =\sum_{t=1}^{T} \left[\sum_{\ell \in\Nc_k}\hat{\thetav}_{[k,t,p]}^{\trasp}\zv_{p}(\xv_{j,t})-  \hat{\thetav}_{[\ell,t,p]}^{\trasp}\zv_{p}(\xv_{j,t})\right]^2 \leq \Oc(\sqrt{T}).
\end{align*} for any learner $k \in \Vc$.
\end{theorem}
\begin{IEEEproof}
The proof is provided in Section~\ref{subsec:proof1}.
\end{IEEEproof}

\begin{theorem}\label{thm2} Under the assumptions {\bf (a1)} - {\bf (a4)}, DOMKL in Algorithm 2 with the parameters $\rho=\eta_l=\eta_g=\Oc(\sqrt{T})$ and kernels $\{\kappa_p: p\in[P]\}$ can achieve the sublinear regrets as
\begin{align*}
    &{\rm regret}_{\rm a}^{k}(T) \\
    &= \sum_{t=1}^{T} \Lc\left(\sum_{p=1}^P \hat{q}_{[k,t,p]}\hat{\thetav}_{[k,t,p]}\zv_{p}(\xv_{k,t}), y_{k,t}\right) \\
    &\;\;\;\;\;\;\;\;\;\;\;\;\;\;\;\;\;\;\;\;\;\;\;\;\;\;\;\;\;\;- \min_{1\leq p\leq P}\sum_{t=1}^{T} \Lc_{[k,t,p]}(\thetav_{[k,p]}^{\star})\leq \Oc(\sqrt{T}),\\
    &{\rm regret}_{\rm d}^{k}(T) \\
    &=\sum_{t=1}^{T} \left[\sum_{\ell \in\Nc_k}\left(\sum_{p=1}^{P}\hat{q}_{[k,t,p]}\hat{\thetav}_{[k,t,p]}^{\trasp}\zv_{p}(\xv_{k,t})\right.\right.\\
    &\;\;\;\;\;\;\;\;\;\;\;\;\;\;\;\;\;\;\;\;\;\;\;\;\;\;\left.\left.-  \sum_{p=1}^{P}\hat{q}_{[\ell,t,p]}\hat{\thetav}_{[\ell,t,p]}^{\trasp}\zv_{p}(\xv_{j,t})\right)   \right]^2 \leq \Oc(\sqrt{T}),
\end{align*} for any learner $k \in \Vc$.
\end{theorem}
\begin{IEEEproof}
The proof is provided in Section~\ref{subsec:proof2}.
\end{IEEEproof}

Theorem~\ref{thm1} and Theorem~\ref{thm2} reveal that DOKL and DOMKL can achieve the optimal sublinear regret bounds when compared with the respective best functions in hindsight. However, it is noticeable that the best functions as to DOMKL and DOKL are from $\bar{\Hc}= \Hc_1 + \cdots + \Hc_P$ and $\Hc_{p} \subseteq \bar{\Hc}$ for a preselected $p \in [P]$, respectively.
In this regards, with a sufficiently large number of kernels, DOMKL can have a potential gain over DOKL.

\vspace{-0.1cm}
\subsection{Proof of Theorem 1}\label{subsec:proof1}

Let $\hat{\thetav}_{[k,t,p]}$ and $\hat{\lambdav}_{[k,t,p]}$ be the output of the proposed DOKL (in Algorithm 1). We first derive the useful lemma:


\begin{lemma}\label{lem2} For any learner $k\in \Vc$, letting $\hat{\gammav}_{[k,t,p]} = \sum_{\ell \in\Nc_k}(\hat{\thetav}_{[k,t,p]}+\hat{\thetav}_{[\ell,t,p]})/2$, we obtain the upper-bound::
\begin{align}
    &\Lc_{[k,t,p]}(\hat{\thetav}_{[k,t+1,p]}) - \Lc_{[k,t,p]} (\thetav_{[k,p]}^{\star})\nonumber\\
    &\leq  \frac{1}{2\rho}( \|\hat{\lambdav}_{[k,t,p]}\|^2-\|\hat{\lambdav}_{[k,t+1,p]}\|^2)-\frac{\rho}{2}\|\hat{\thetav}_{[k,t+1,p]}-\hat{\gammav}_{[k,t,p]}\|^2\nonumber\\
    &+\frac{\eta_l}{2}(\|\hat{\thetav}_{[k,t,p]}-\thetav_{[k,p]}^{\star}\|^{2} - \|\hat{\thetav}_{[k,t+1,p]}-\thetav_{[k,p]}^{\star}\|^{2})\nonumber\\
    &-\frac{\eta_l}{2}(\|\hat{\thetav}_{[k,t+1,p]})-\hat{\thetav}_{[k,t,p]}\|^2)\nonumber\\
    &+\frac{\rho}{2}(\|\hat{\gammav}_{[k,t,p]}-\thetav_{[k,p]}^{\star}\|^2 - \|\hat{\gammav}_{[k,t+1,p]}-\thetav_{[k,p]}^{\star}\|^2).\label{eq:proof0}
\end{align} 
\end{lemma}
\begin{IEEEproof} The proof is provided in Appendix~\ref{app:lem2}.
\end{IEEEproof}

We are now ready to prove Theorem~\ref{thm1}.

{\em (a) The proof of learning accuracy:} From the convexity of the loss function, we obtain the following inequality:
\begin{align}
    &\Lc_{[k,t,p]} (\hat{\thetav}_{[k,t,p]}) - \Lc_{[k,t,p]} (\hat{\thetav}_{[k,t+1,p]})\nonumber\\
    &\leq \left\langle \nabla\Lc_{[k,t,p]}(\hat{\thetav}_{[k,t,p]}),\; \hat{\thetav}_{[k,t,p]} - \hat{\thetav}_{[k,t+1,p]} \right\rangle \nonumber\\
    &= \left\langle \frac{1}{\sqrt{ \eta_l}}\nabla\Lc_{[k,t,p]}(\hat{\thetav}_{[k,t,p]}),\; \sqrt{\eta_l}(\hat{\thetav}_{[k,t,p]} - \hat{\thetav}_{[k,t+1,p]}) \right\rangle\nonumber\\
    & \stackrel{(a)}{\leq} \frac{1}{2 \eta_l} \|\nabla\Lc_{[k,t,p]}(\hat{\thetav}_{[k,t,p]})\|^2 + \frac{\eta_l}{2}\|\hat{\thetav}_{[k,t,p]} - \hat{\thetav}_{[k,t+1,p]}\|^2,\label{eq:proof5}
\end{align} where (a) is due to the Fenchel-Young inequality \cite{rockafellar1970convex}. From (\ref{eq:proof0}) and (\ref{eq:proof5}), we can get:
\begin{align}
    &\Lc_{[k,t,p]}(\hat{\thetav}_{[k,t,p]}) - \Lc_{[k,t,p]}(\thetav_{[k,p]}^{\star}) \nonumber\\
    &\leq  \frac{1}{2\rho} (\|\hat{\lambdav}_{[k,t,p]}\|^2-\|\hat{\lambdav}_{[k,t+1,p]}\|^2)+\frac{1}{2\eta_l}\|\nabla\Lc_{[k,t,p]}(\hat{\thetav}_{[k,t,p]})\|^2\nonumber\\
    &+\frac{\rho}{2}(\|\hat{\gammav}_{[k,t,p]}-\thetav_{[k,p]}^{\star}\|^2 - \|\hat{\gammav}_{[k,t+1,p]}-\thetav_{[k,p]}^{\star}\|^2)\nonumber\\
    &+\frac{\eta_l}{2}(\|\hat{\thetav}_{[k,t,p]}-\thetav_{[k,p]}^{\star}\|^{2} - \|\hat{\thetav}_{[k,t+1,p]}-\thetav_{[k,p]}^{\star}\|^{2}).\nonumber
\end{align} From the above inequality and the telescoping sum, we have:
\begin{align}
 &\sum_{t=1}^{T} \Lc_{[k,t,p]} (\hat{\thetav}_{[k,t,p]}) - \Lc_{[k,t,p]}(\thetav_{[k,p]}^{\star})\nonumber\\
 &\leq \frac{1}{2\rho} (\|\hat{\lambdav}_{[k,1,p]}\|^2-\|\hat{\lambdav}_{[k,T+1,p]}\|^2)\nonumber\\
     &+\frac{\rho}{2}(\|\hat{\gammav}_{[k,1,p]}-\thetav_{[k,p]}^{\star}\|^2 - \|\hat{\gammav}_{[k,T+1,p]}-\thetav_{[k,p]}^{\star}\|^2)\nonumber\\
     &+\frac{1}{2 \eta_l}\sum_{t=1}^{T}\|\nabla\Lc_{[k,t,p]}(\hat{\thetav}_{[k,t,p]})\|^2\nonumber\\
     &\stackrel{(a)}{\leq} \frac{\rho}{2}\|\hat{\gammav}_{[k,1,p]}-\thetav_{[k,p]}^{\star}\|^2 + \frac{1}{2 \eta_l}\sum_{t=1}^{T}\|\nabla\Lc_{[k,t,p]}(\hat{\thetav}_{[k,t,p]})\|^2,\label{eq:final}
\end{align} where (a) is due to the fact $\hat{\lambdav}_{[k,1,p]} = \zerov$, $\|\hat{\lambdav}_{[k,T+1,p]}\|^2\geq0$, and $\|\hat{\gammav}_{[k,T+1,p]}-\thetav_{[k,p]}^{\star}\|^2\geq 0$. Finally, from the assumptions {\bf (a1)} and {\bf (a2)}, we can get:
\begin{align}
    \sum_{t=1}^{T}\Lc_{[k,t,p]} (\hat{\thetav}_{[k,t,p]}) - \Lc_{[k,t,p]}(\thetav_{[k,p]}^{\star}) \leq \frac{\rho C}{2} + \frac{T G}{2\eta_l}.
\end{align} Setting $\rho=\eta_l=\Oc(\sqrt{T})$, we can verify that DOKL (in Algorithm 1) achieves the sublinear regret, which completes the proof of the part (a).

{\em (b) The proof of consensus violation:} We first obtain the following upper-bound:
\begin{align}
    &\left[\sum_{\ell \in \Nc_k}\left( \hat{\thetav}_{[k,t,p]}^{\trasp}\zv_{p}(\xv_{k,t}) - \hat{\thetav}_{[\ell,t,p]}^{\trasp}\zv_{p}(\xv_{k,t}) \right)\right]^2\nonumber\\
    & \stackrel{(a)}{\leq}\left\|\sum_{\ell\in\Nc_{k}} \hat{\thetav}_{[k,t,p]}-\hat{\thetav}_{[\ell,t,p]}\right\|^2\Big\|\zv_{p}(\xv_{k,t})\Big\|^2\nonumber\\
    &\stackrel{(b)}{\leq} \left\|\sum_{\ell \in\Nc_{k}} \hat{\thetav}_{[k,t,p]}-\hat{\thetav}_{[\ell,t,p]}\right\|^2,\label{eq:regret_d1}
\end{align} where (a) follows the Cauchy-Schwartz inequality and (b) is due to the fact that $\|\zv_p(\xv_{k,t})\|_2^2\leq 1$ from (\ref{eq:zv_o}). Also, the following inequality is obtained:
\begin{align}
    &\left\|\sum_{\ell \in\Nc_{k}} \hat{\thetav}_{[k,t,p]}-\hat{\thetav}_{[\ell,t,p]}\right\|^2 \nonumber\\
    &\stackrel{(a)}{=} \frac{4}{\rho^2}\left\| \hat{\lambdav}_{[k,t,p]}-\hat{\lambdav}_{[k,t-1,p]}\right\|^2=4\left\|\hat{\thetav}_{[k,t,p]}-\hat{\gammav}_{[k,t,p]}\right\|^2 \nonumber\\
    &\stackrel{(b)}{\leq} 4\left\|\hat{\thetav}_{[k,t+1,p]} - \hat{\gammav}_{[k,t,p]}\right\|^2\nonumber\\
    &\stackrel{(c)}{\leq} \frac{8B}{\rho} + \frac{4}{\rho^2} (\|\hat{\lambdav}_{[k,t,p]}\|^2-\|\hat{\lambdav}_{[k,t+1,p]}\|^2)\nonumber\\
    &+4(\|-\thetav_{[k,p]}^{\star}+\hat{\gammav}_{[k,t,p]}\|^2 - \|-\thetav_{[k,p]}^{\star}+\hat{\gammav}_{[k,t+1,p]}\|^2)\nonumber\\
    &+\frac{2\eta_l}{\rho}(\|\hat{\thetav}_{[k,t,p]}-\thetav_{[k,p]}^{\star}\|^{2} - \|\hat{\thetav}_{[k,t+1,p]}-\thetav_{[k,p]}^{\star}\|^{2}),\label{eq:regret_d2}
\end{align} where (a) is due to the fact that $\hat{\thetav}_{[k,t+1,p]}-\hat{\gammav}_{[k,t+1,p]}=\frac{1}{\rho}(\hat{\lambdav}_{[k,t,p]} - \hat{\lambdav}_{[k,t+1,p]})$, (b) follows the \cite[Lemma 3]{wang2013online}, and (c) is from the rearrangement of (\ref{eq:proof0}) in Lemma~\ref{lem2} and using the assumption {\bf (a3)}. From (\ref{eq:regret_d1}) and (\ref{eq:regret_d2}), and using the telescoping sum, we can get:
\begin{align}
&{\rm Regret}_{\rm d}^k (T)=\sum_{t=1}^{T} \left[\sum_{\ell \in \Nc_k}\hat{f}_{[k,t,p]}(\xv_{k,t}) -\hat{f}_{[\ell,t,p]}(\xv_{k,t}) \right]^2\nonumber\\
    &\leq \sum_{t=1}^{T}\left\|\sum_{\ell \in\Nc_{k}} \hat{\thetav}_{[k,t+1,p]}-\hat{\thetav}_{[\ell,t+1,p]}\right\|^2\nonumber\\
    &\leq \frac{8B T}{\rho}+ 4\|\hat{\gammav}_{[k,1,p]}-\thetav_{[k,p]}^{\star}\|^2+\frac{2\eta_l}{\rho}\|\hat{\thetav}_{[k,1,p]}-\thetav_{[k,p]}^{\star}\|^2\nonumber\\
    &\leq \frac{8B T}{\rho}+4C + \frac{4\eta_l}{\rho}C.\label{eq:thm1-3}
\end{align} Setting $\rho=\eta_l=\Oc(\sqrt{T})$, the sublinear regret is achieved, which completes the proof.

\subsection{Proof of Theorem 2}\label{subsec:proof2}

We prove the sublinear regrets of DOMKL in Algorithm 2.

\vspace{0.2cm}
{\em (a) The proof of learning accuracy:} 
We first give the key lemma:
\vspace{0.1cm}
\begin{lemma}\label{lem3} Setting $\eta_g = \Oc(\sqrt{T})$ and using the weights in (\ref{eq:weight-update}), the following sublinear regret is achieved:
\begin{align}
        &\sum_{t=1}^{T} \Lc\left(\sum_{p=1}^{P}\hat{q}_{[k,t,p]}\hat{f}_{[k,t,p]}(\xv_{k,t}),y_{k,t}\right) \nonumber\\
        &\;\;\;\;\;\;\;\;\;\;\;\;\;\;\;\; - \min_{1\leq p \leq P} \sum_{t=1}^T \Lc_{[k,t,p]}(\hat{\thetav}_{[k,t+1,p]}) \leq \Oc(\sqrt{T}),\label{eq:M_proof2}
    \end{align} for any learner $k \in \Vc$.
\end{lemma}
\begin{IEEEproof} The proof is provided in Appendix~\ref{app:lem3}.
\end{IEEEproof}
From Theorem~\ref{thm1}, we know that for any kernel $p \in [P]$, 
\begin{align}
        \sum_{t=1}^{T} \Lc_{[k,t,p]}(\hat{\thetav}_{[k,t+1,p]}) - \Lc_{[k,t,p]}(\thetav_{[k,p]}^{\star}) \leq \Oc(\sqrt{T}).\label{eq:M_proof1}
\end{align} By integrating (\ref{eq:M_proof1}) and (\ref{eq:M_proof2}), the proof is completed.

{\em (b) The proof of consensus violation:}  We first obtain the following upper-bound on the discrepancy at time $t$:
\begin{align}
&\left[\sum_{\ell \in\Nc_k} \left( \sum_{p=1}^{P}\hat{q}_{[k,t,p]}\hat{f}_{[k,t,p]}(\xv_{k,t}) - \sum_{p=1}^{P}\hat{q}_{[\ell,t,p]}\hat{f}_{[\ell,t,p]}(\xv_{k,t})\right) \right]^2\nonumber\\
&\stackrel{(a)}{\leq} 2\sum_{p=1}^{P} \hat{q}_{[k,t,p]}^2 \left[\sum_{\ell\in\Nc_k} \hat{\thetav}_{[k,t,p]}^{\trasp}\zv_{p}(\xv_{j,t}) -   \hat{\thetav}_{[\ell,t,p]}^{\trasp}\zv_{p}(\xv_{j,t}) \right]^2\nonumber\\
&+2D_t\nonumber\\
&\leq 2\sum_{p=1}^{P} \hat{q}_{[k,t,p]}^2 \left\|\sum_{\ell\in\Nc_{k}} \hat{\thetav}_{[k,t,p]}-\hat{\thetav}_{[\ell,t,p]}\right\|^2+2D_t,\label{eq:proof-thm2-1}
\end{align} where (a) is due to the fact that $(A+B)^2\leq 2(A^2+B^2)$ and 
\begin{align}
D_t = \left[\sum_{\ell \in \Nc_k} \sum_{p=1}^{P}\Big(\hat{q}_{[k,t,p]} - \hat{q}_{[\ell,t,p]}\Big)\hat{\thetav}_{[\ell,t,p]}\zv_p(\xv_{k,t}) \right]^2.
\end{align}
From (\ref{eq:proof-thm2-1}), we obtain the following upper-bound:
\begin{align}
    &{\rm Regret}_{d}^k(T) \nonumber\\
    &\;\; \leq 2\sum_{p=1}^{P} \hat{q}_{[k,t,p]}^2 \sum_{t=1}^{T}\left\|\sum_{\ell \in\Nc_{k}} \hat{\thetav}_{[k,t,p]}-\hat{\thetav}_{[\ell,t,p]}\right\|^2+2\sum_{t=1}^TD_t\nonumber\\ 
    &\;\; \stackrel{(a)}{\leq} 2\sum_{p=1}^{P} \hat{q}_{[k,t,p]}^2 \left(\frac{8B T}{\rho}+4C+ \frac{4\eta_l}{\rho}C\right)+2\sum_{t=1}^TD^t\nonumber\\
    &\;\; \stackrel{(b)}{\leq} \frac{16 B T}{\rho}+8C + \frac{8\eta_l}{\rho}C+2\sum_{t=1}^TD_t,\label{eq:proof10}
\end{align} where (a) is from (\ref{eq:thm1-3}) and (b) is due to the fact that $\sum_{p=1}^{P}\hat{q}_{[k,t,p]}^2\leq 1$. Also, we have that
\begin{align}
    \sum_{t=1}^T D_t &\leq \sum_{t=1}^{T} \epsilon_{t}^2  \sum_{p=1}^{P}\sum_{\ell\in \Nc_k} \left\|\hat{\thetav}_{[\ell,t,p]}\right\|^2\nonumber\\
    &\stackrel{(a)}{\leq} |\Nc_k|C P\sum_{t=1}^{T}\epsilon_t^2\stackrel{(b)}{\leq} \Oc(\sqrt{T}),\label{eq:proof11}
\end{align} where (a) is from the assumption {\bf (a2)} and (b) is from the assumption {\bf (a4)}. From (\ref{eq:proof10}) and (\ref{eq:proof11}), and setting $\rho=\eta_l=\Oc(\sqrt{T})$, the sublinear regret of discrepancy is achieved, which completes the proof.


\section{Experiments}\label{sec:Exp}

In this section, we verify the effectiveness of the proposed DOMKL via experiments on various online regression and time-series prediction tasks with real datasets. { In particular, the superiority of DOMKL will be demonstrated by showing stable performances on various online learning tasks and network structures (e.g., network connectivity and size), and the great advantage of using multiple kernels.} A least-square loss function, which has been widely used in the above learning tasks \cite{shen2019random, hong2020active, bouboulis2017online}, is assumed:
\begin{equation}
    \Lc(\thetav^{\trasp}\zv_{p}(\xv_{t}), y_{t})= (y_{t} - \thetav^{\trasp}\zv_{p}(\xv_{t}))^2.\label{loss-fn}
\end{equation} { For the sake of comparison, we construct a centralized approach based on OMKL in \cite{shen2019random, hong2020active}, which is named COMKL. In this method, at time $t$, every learner $k \in \Vc$ transmits its local data $(\xv_{k,t}, y_{k,t})$ (or the associated local gradient $\nabla\Lc(\hat{\thetav}_{[t,p]}^{\trasp}\zv_p(\xv_{k,t}),y_{k,t})$) to a central server. Using them, the central server learns a common function 
$\hat{f}_{t+1}(\xv) = \sum_{p=1}^{P} \hat{q}_{[t+1,p]}\hat{f}_{[t+1,p]}(\xv)$ via \cite[Algorithm 1]{hong2020active} with a modified local step. This modification is required as the central server observes the $K$ labeled data at every time, whereas in OMKL \cite{shen2019random, hong2020active}, one labeled data arrives. Then, the central server broadcasts the estimated function $\hat{f}_{t+1}(\cdot)$ to all $K$ nodes. This method can be regarded as a {\em centralized} federated learning based on multiple kernels.  Obviously, COMKL reduces to OMKL when $K=1$. Specifically, in the modified local step, the parameter of each kernel function $\hat{f}_{[t+1,p]}$ (parameterized by $\hat{\thetav}_{[t+1,p]}$) is updated via mini-batch OGD with the batch size $K$:
\begin{equation}
    \hat{\thetav}_{[t+1,p]} = \hat{\thetav}_{[t,p]} -  \frac{\eta_l}{K}\sum_{k=1}^{K}\nabla\Lc(\hat{\thetav}_{[t,p]}^{\trasp}\zv_p(\xv_{k,t}),y_{k,t}).\label{eq:modified_local}
\end{equation} We remark that the learning accuracy of COMKL can be regarded as the performance limit (or the lower bound) of all possible multiple kernel-based distributed online learning algorithms (including our DOMKL). Definitely the other kernel-based learning algorithms (e.g., online multiple kernel regression \cite{sahoo2014online}, online multiple kernel learning with a budget \cite{kivinen2004online}, and so on) can be incorporated into the above centralized framework. However, they are excluded for comparisons since their performances are much worse than COMKL, as expected from the numerical results in \cite{hong2020active}  and references therein. Also, we consider the state-of-the-art kernel-based distributed online learning algorithm (named RFF-DOKL) \cite{bouboulis2017online}. As far as we know, RFF-DOKL is the only kernel-based algorithm suitable for our distributed online learning framework with continuous streaming data. As explained in Section~\ref{sec:intro}, the other kernel-based distributed learning algorithms were not customized in online learning, thus being lack of scalability with respect to the number of incoming data.} In summary, the following online learning algorithms will be used for our experiments:
\begin{itemize}
    \item {\bf COMKL}: The centralized online multiple kernel-based learning algorithm, which is the variant of OMKL \cite{shen2019random, hong2020active} with the modified local step in \eqref{eq:modified_local}. Note that the underlying optimization techniques are based on OGD and a Hedge algorithm.
     \item {\bf RFF-DOKL}: The distributed online single kernel-based learning algorithm, which is developed based on OGD and a diffusion strategy \cite{bouboulis2017online}. (Gaussian kernel with $\sigma^2$).
    \item {\bf DOKL} : The proposed distributed online single kernel-based learning algorithm, which is based on online ADMM and a distributed Hedge algorithm. (Gaussian kernel with $\sigma^2$).
     \item {\bf DOMKL}: The proposed distributed online {\it multiple} kernel-based learning algorithm, where the kernel dictionary consists of the $17$ Gaussian kernels (i.e., $P=17$) with the parameters (called bandwidths) 
\begin{equation}\label{eq:GK}
    \sigma_{p}^{2}=10^{\frac{p-9}{2}},~ p=1,\dots,17.
\end{equation}
\end{itemize} For RFF-DOKL and DOKL, the bandwidth of a Gaussian kernel (i.e, $\sigma^2$) will be specified when they are used.
%
%
\begin{table}[t]
{\caption{Performance comparisons of MSE ($\times 10^{-2}$) and CV ($\times 10^{-2}$) as a function of the parameters 
$\eta_g$ and $\rho$. ($K=10$ and $\alpha_c=0.25$). }}\vspace{-0.2cm}
\vspace{0.1cm}
\begin{center}
\begin{tabular}{cccccc}
\toprule
\multirow{2}{*}{$\eta_{g}$} & \multirow{2}{*}{$\rho$} &  \multicolumn{2}{c}{Conductivity} & \multicolumn{2}{c}{Weather}  \\ \cline{3-6}
&  & {\rm MSE} & {\rm CV} & {\rm MSE} &  {\rm CV}\\\hline
10 & 10 & 5.23 & 3.11 & 0.41 & 0.25 \\
\cellcolor{red!15}10 & \cellcolor{red!15}100 &\cellcolor{red!15}4.78 &\cellcolor{red!15}1.10 &\cellcolor{red!15}0.49 &\cellcolor{red!15}0.19 \\
10 & 1000 & 4.98 & 0.46 & 0.86 & 0.21\\\hline
100 & 10 & 5.68 & 2.77 & 0.55 & 0.28 \\
100 & 100 & 4.93 & 0.95 & 0.66 & 0.23 \\
100 & 1000 & 5.27 & 0.46 & 1.16 & 0.25\\
\bottomrule
\end{tabular}
\end{center}
\label{table:parameter}
\end{table}


{ We investigate the performances of the above algorithms in terms of  {\em learner-wise} and {\em network-wise} aspects. They are measured by learning accuracy and consensus violation, respectively. Recall that $\hat{y}_{k,t} = \hat{f}_{[k,t]}(\xv_{k,t})$ and $y_{k,t}$ stand for an estimated label and a true label at time $t$ in the learner $k$, where $\hat{f}_{[k,t]}$ is generated at time $t-1$ via Algorithm 1 (resp. Algorithm 2) for the proposed DOKL (resp. DOMKL). Through the experiments, it is assumed that communication graphs in our experiments are connected. This assumption does not violate the generality as the proposed algorithms can be straightforwardly applied to each connected component in parallel if a communication graph is not connected.

{\em i) Learning accuracy:} We evaluate the accuracy of a learned function with respect to the associated local data, which is measured by the mean-square-error (MSE) as
\begin{equation*}
\mbox{MSE}(t)= \frac{1}{tK}\sum_{\tau=1}^{t}\sum_{k=1}^{K}(\hat{f}_{[k,\tau]}(\xv_{k,\tau})-y_{k,\tau})^{2},
\end{equation*} with the initial value $\mbox{MSE}(1)=1$. It is well matched to our loss function in \eqref{loss-fn}. However, this metric alone is not sufficient as it could not reflect a network-wise performance (i.e., the closeness of learned functions). We address this problem by introducing the following second metric.

{\em ii) Consensus violation:} Since a communication graph is assumed to be connected, every learner in the network aims at learning a common function. By taking this into account,  the network-wise consensus is measured as 
\begin{align*}
    &\mbox{CV}(t)=\nonumber\\
    &\frac{1}{tK(K-1)}\sum_{\tau=1}^{t}\sum_{k=1}^{K}\sum_{\ell=1: \ell\neq k}^{K} (\hat{f}_{[k,\tau]}(\xv_{k,\tau}) -\hat{f}_{[\ell,\tau]}(\xv_{k,\tau}))^2,
\end{align*} with the initial value $\mbox{CV}(1)=1$.
}

To capture the impact of network structures, in our experiments, we consider a {\em random} communication graph consisting of $K$ nodes (or learners). This graph is constructed by starting with a set of $K$ isolated nodes and adding successive edges between them randomly according to a {\em connection} probability $\alpha_c \in (0,1]$. Consequently, every learner in the network can have $\alpha_c K$ neighboring learners in average. Due to the randomness of the learning algorithms (e.g., COMKL, RFF-DOKL, DOKL, and DOMKL) and a communication graph, the {\em averaged} MSE and CV performances over $500$ trials are evaluated. On constructing a random communication graph, it might be possible that the resulting graph is not connected. This sample is removed (i.e., not counted in $500$ trials). In the proposed DOKL and DOMKL (i.e., Algorithm 1 and Algorithm 2), the following hyper-parameters are used:
\begin{equation}\label{hyper}
    \rho = 100,\; \eta_l = 10,\; \eta_g = 10,\; \text{and}~ M=50.
\end{equation} In fact, $\eta_l$ has little impact on the performances, whereas $\rho$ and $\eta_g$ can control the tradeoff between learning accuracy and consensus violation, as provided in Table I. For instance, if $\rho$ is large, the network tends to build a strong consensus at the expense of accuracy-performance losses at learners. Unfortunately, the optimal choices of $\rho$ and $\eta_g$ rely on the both target MSE and CV performances, and datasets. In real-world online learning tasks, it might not be tractable to optimize these parameters because of the sequential nature of incoming data. Such optimization is an interesting open problem even in the simpler centralized setting  \cite{hong2020active}, which is beyond the scope of this paper. In all datasets, we used the hyper-parameters in \eqref{hyper}, which is determined on the basis of the results in Table I. 
The datasets to be used for the experiments are given in Section~\ref{subsec:dataset}, and the experimental results and some discussions are provided in Section~\ref{subsec:discuss}.


\begin{table}\label{tb:data}
{\caption{Summary of Real Datasets for Experiments}}
\vspace{-0.2cm}
\label{tb:DataSummary}
\centering
\begin{tabular}{ c||c|c|c } 
\hline
\multicolumn{4}{c}{\cellcolor{lightgray}\bf Regression task} \\ \hline
Datasets & \# of features & \# of data & feature type  \\
\hline
Weather  & 21 & 7750 & real  \\ 
\hline
Conductivity  & 81 & 11000 & real \\ 
\hline
Wave energy  & 48  & 9500 & real \\ 
\hline
Twitter & 77 & 98704 & real \& integer\\
\hline
\multicolumn{4}{c}{\cellcolor{lightgray}\bf Time series prediction task} \\ \hline
Datasets & \# of features & \# of data & feature type  \\
\hline
Air quality  & 13 & 7322 & real \\ 
\hline
Traffic  & 5,10 & 6500 & real  \\ 
\hline
Temperature  & 5,10 & 5500 & real \\ 
\hline
\end{tabular}
\end{table}

\subsection{Descriptions of online learning tasks and datasets}\label{subsec:dataset}

We describe online regression and time-series prediction tasks and  real-world datasets for the experiments.

\subsubsection{Online regression tasks}
For these tasks, we consider the following real datasets from UCI Machine Learning Repository, which are also summarized in Table II:
\begin{itemize}
    \item {\bf Weather} \cite{D.Cho2020} : The dataset contains 7750 samples obtained from LDAPS model operated by the Korea Meteorological Administration during 2015-2017, where the features show the geographical variables of Seoul. The goal is to predict the minimum temperature of next day. 
    \item {\bf Conductivity} \cite{K.Hamidieh2018} : The dataset contains 11000 samples of extracted from superconductors, where each feature in $\xv_{t} \in \RR^{81}$ represents critical information to construct superconductor such as density and mass of atoms. The purpose is to predict the critical temperature which creates superconductor.
    \item {\bf Wave energy} \cite{M.Neshat2018} : This dataset contains 9500 samples consisted of positions and absorbed power obtained from wave energy converters (WECs) in four real wave scenarios from the southern coast of Australia. The objective is to predict the total power energy of the farm.
    \item {\bf Twitter \cite{Kawala2013}:} This dataset contains 98704 samples consisted of buzz events from Twitter, where each attributes are used to predict the popularity of a topic. Higher value indicates more popularity. 
\end{itemize} { To build a distributed learning framework, the above data samples are distributed to the $K$ nodes in the following way. The entire data samples are partitioned into the $K$ parts of the equal size 
$T=\left\lfloor \mbox{the total number of data}/K\right\rfloor$, where some remaining data samples can be excluded. The resulting partition is denoted as
$\{\Dc_{1},...,\Dc_{K}\}$, where $|\Dc_{k}|=T, \forall K \in [K]$ and $\Dc_{k}\cap\Dc_{\ell}=\phi$ for any $k, \ell \in \Vc$ with $k\neq \ell$. Then, $\Dc_{k}=\{(\xv_{k,t}, y_{k,t}): t=1,...,T\}$ is used as the local dataset of the learner $k$ for $k \in \Vc$. At time $t$, each learner $k$ receives the data $(\xv_{k,t}, y_{k,t})$ from $\Dc_k$. Note that although there are lots of possible partitions of the equal size $T$, they do not affect the performances of the algorithms considered in the experiments. In contrast, it is not true for time-series prediction tasks in the next subsection.}

\begin{figure*}[!h]
\centering
\subfigure[Conductivity data]{
\includegraphics[width=0.34\linewidth]{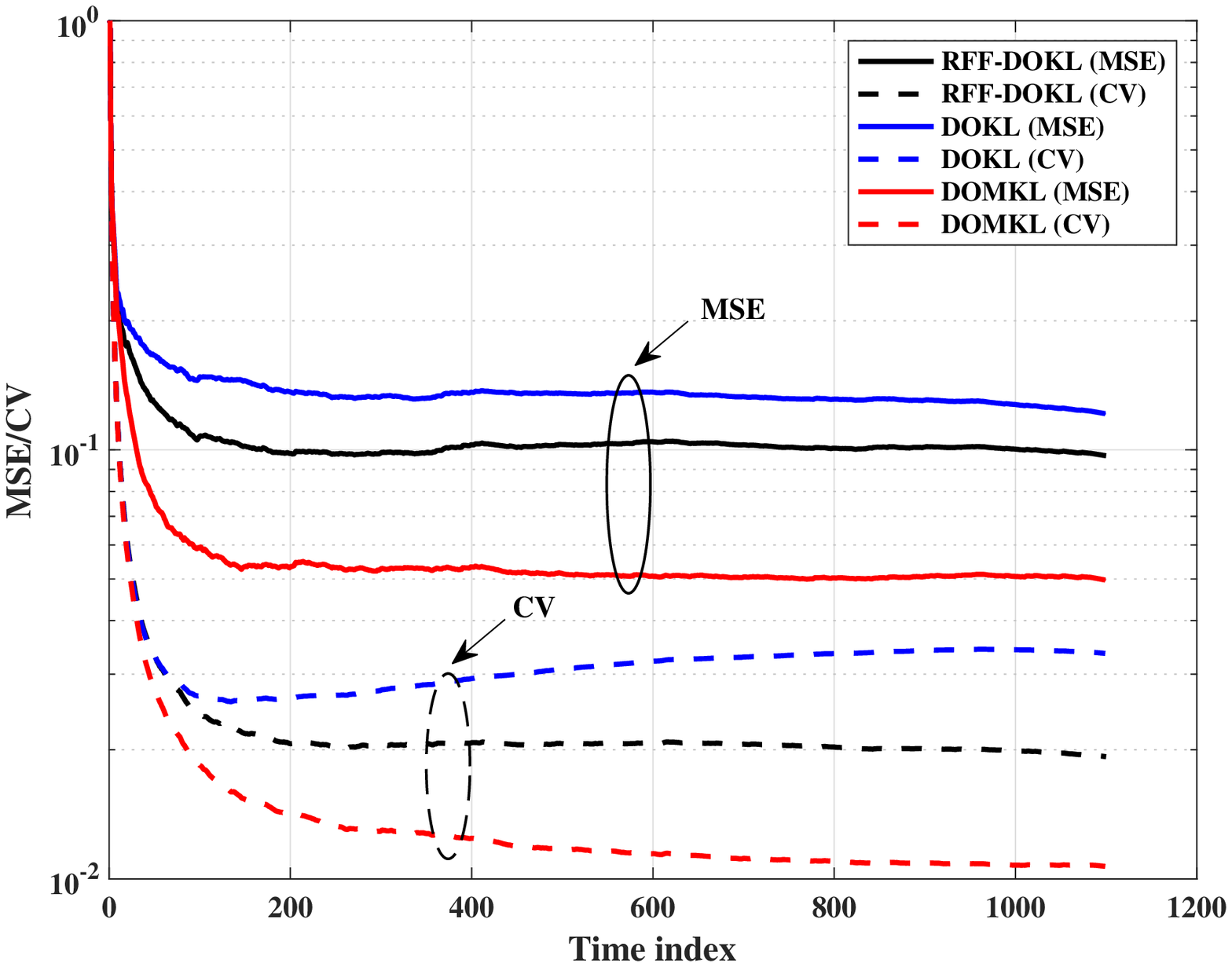}
}
\hspace{-0.7cm}\centering
\subfigure[Weather data]{
\includegraphics[width=0.34\linewidth]{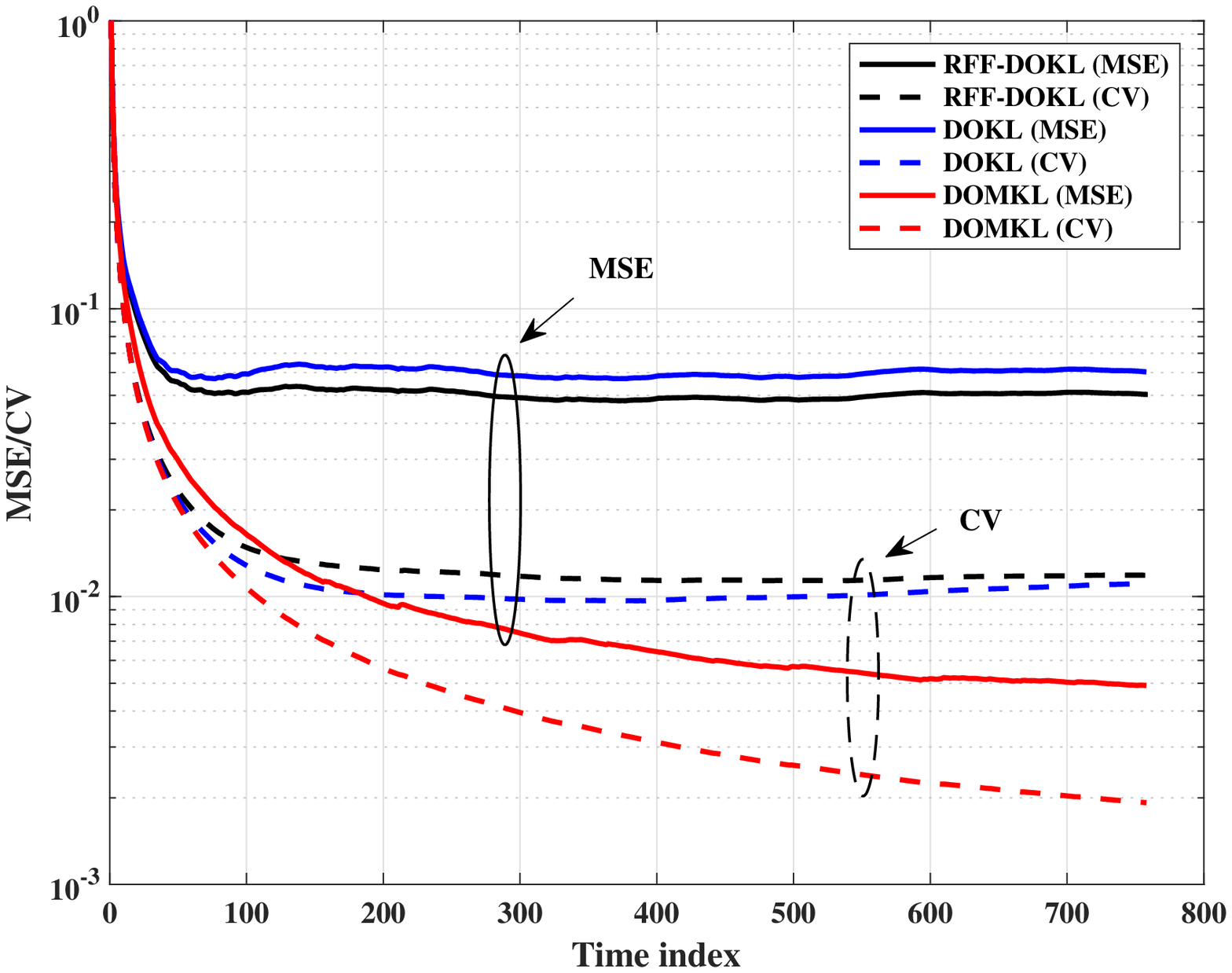}
}
\hspace{-0.7cm}\centering
\subfigure[Wave energy data]{
\includegraphics[width=0.34\linewidth]{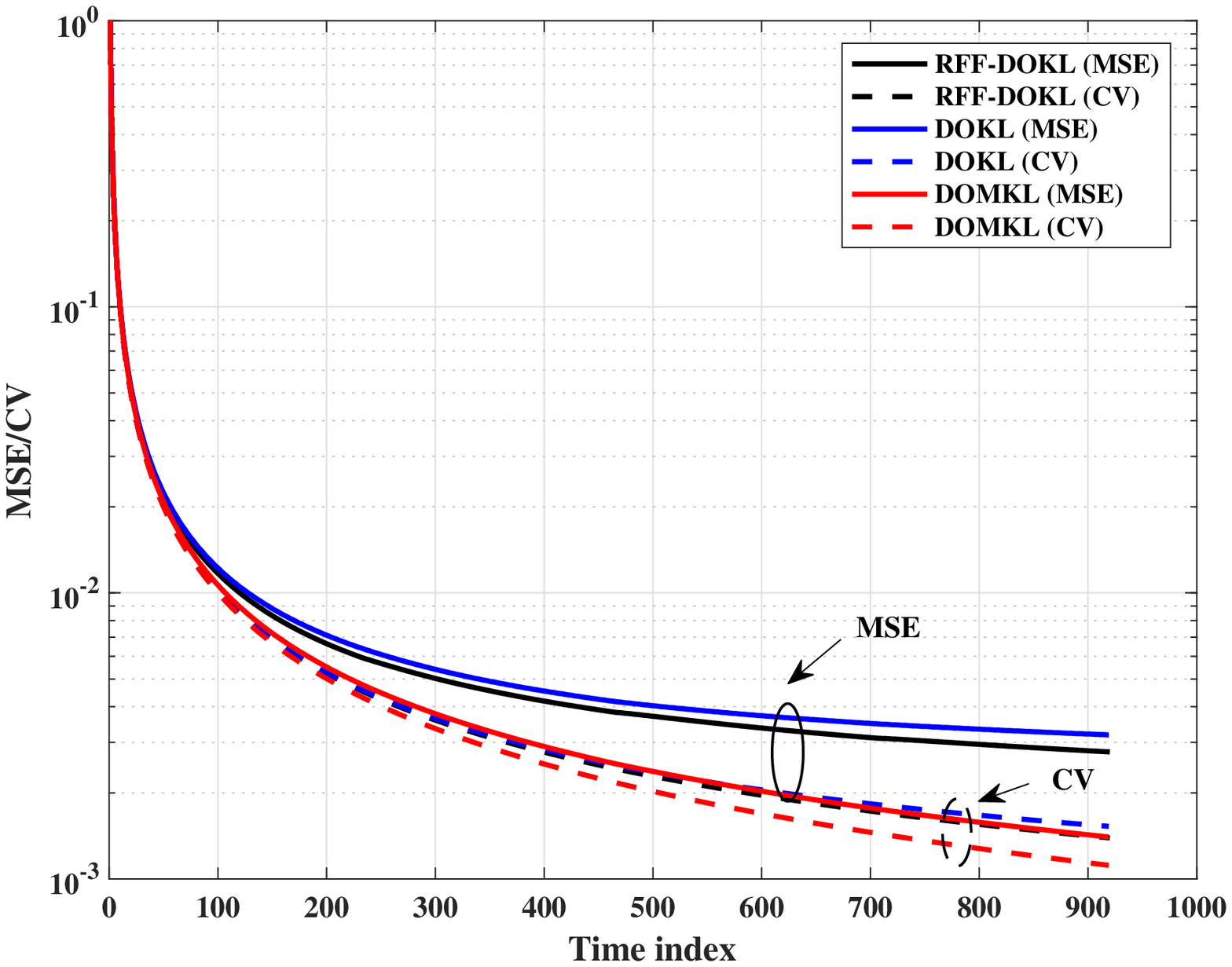}
}
\hspace{-0.7cm}\centering
\subfigure[Air quality data]{
\includegraphics[width=0.34\linewidth]{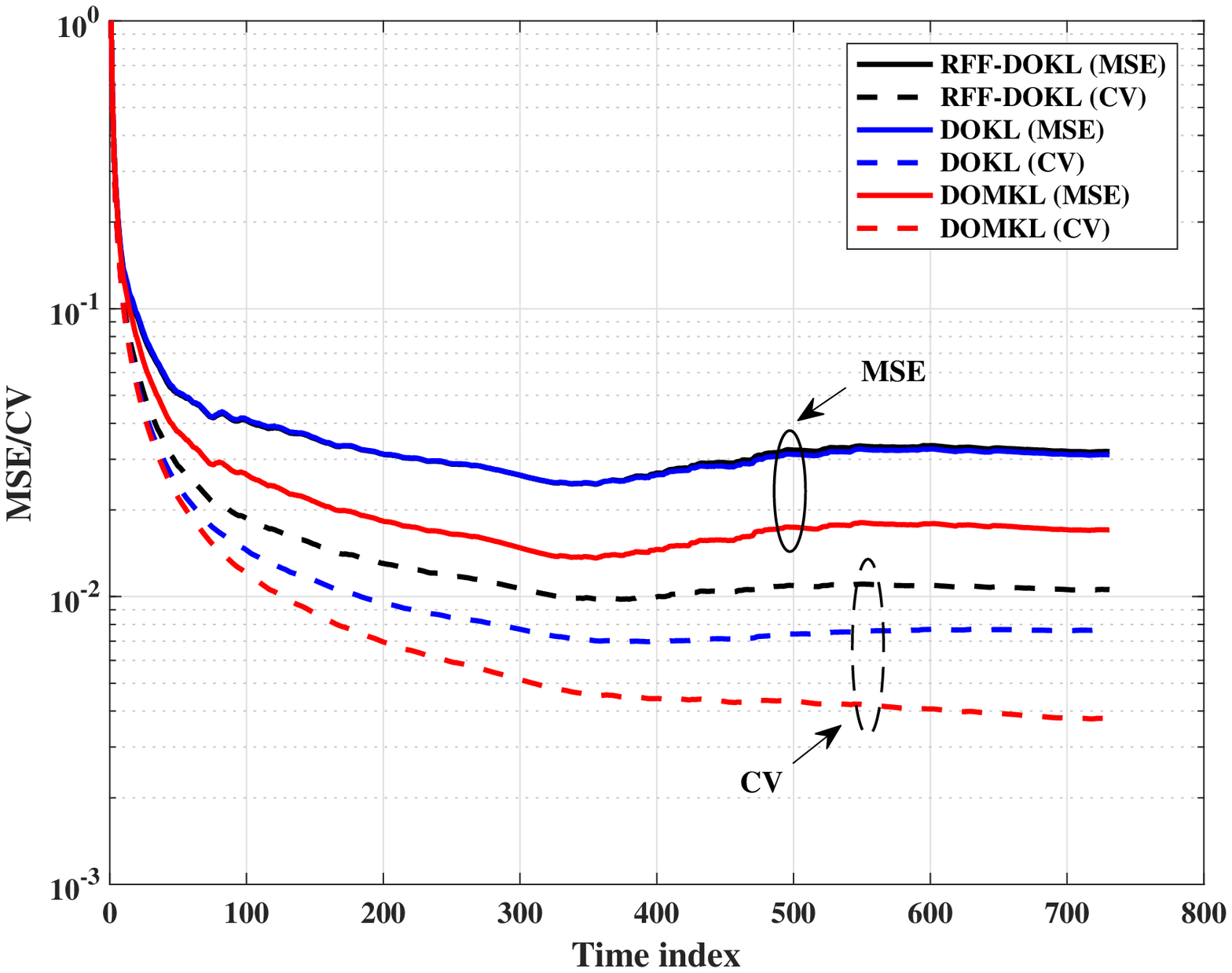}
}
\hspace{-0.7cm}\centering
\subfigure[Traffic data]{
\includegraphics[width=0.34\linewidth]{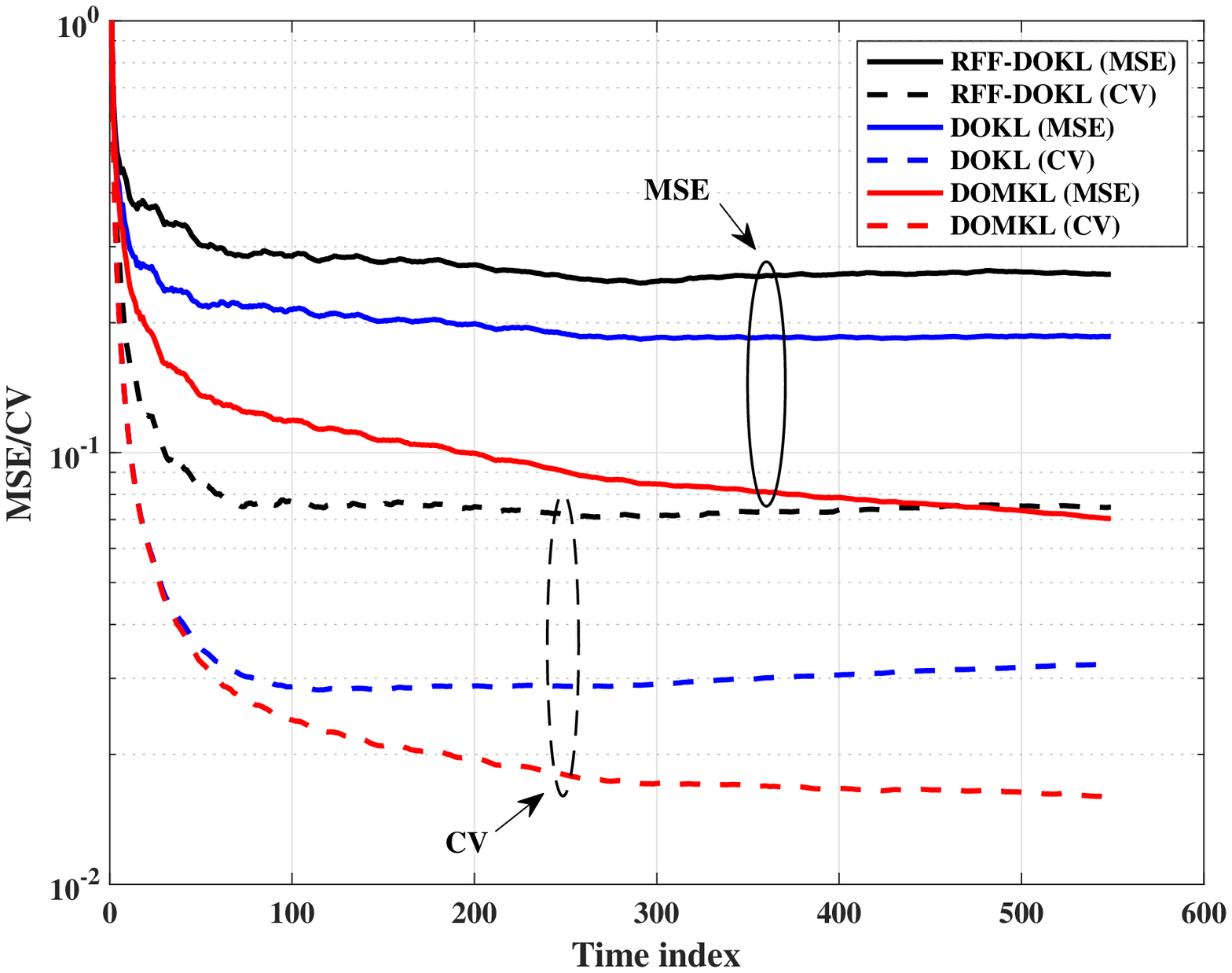}
}
\hspace{-0.7cm}\centering
\subfigure[Temperature data]{
\includegraphics[width=0.34\linewidth]{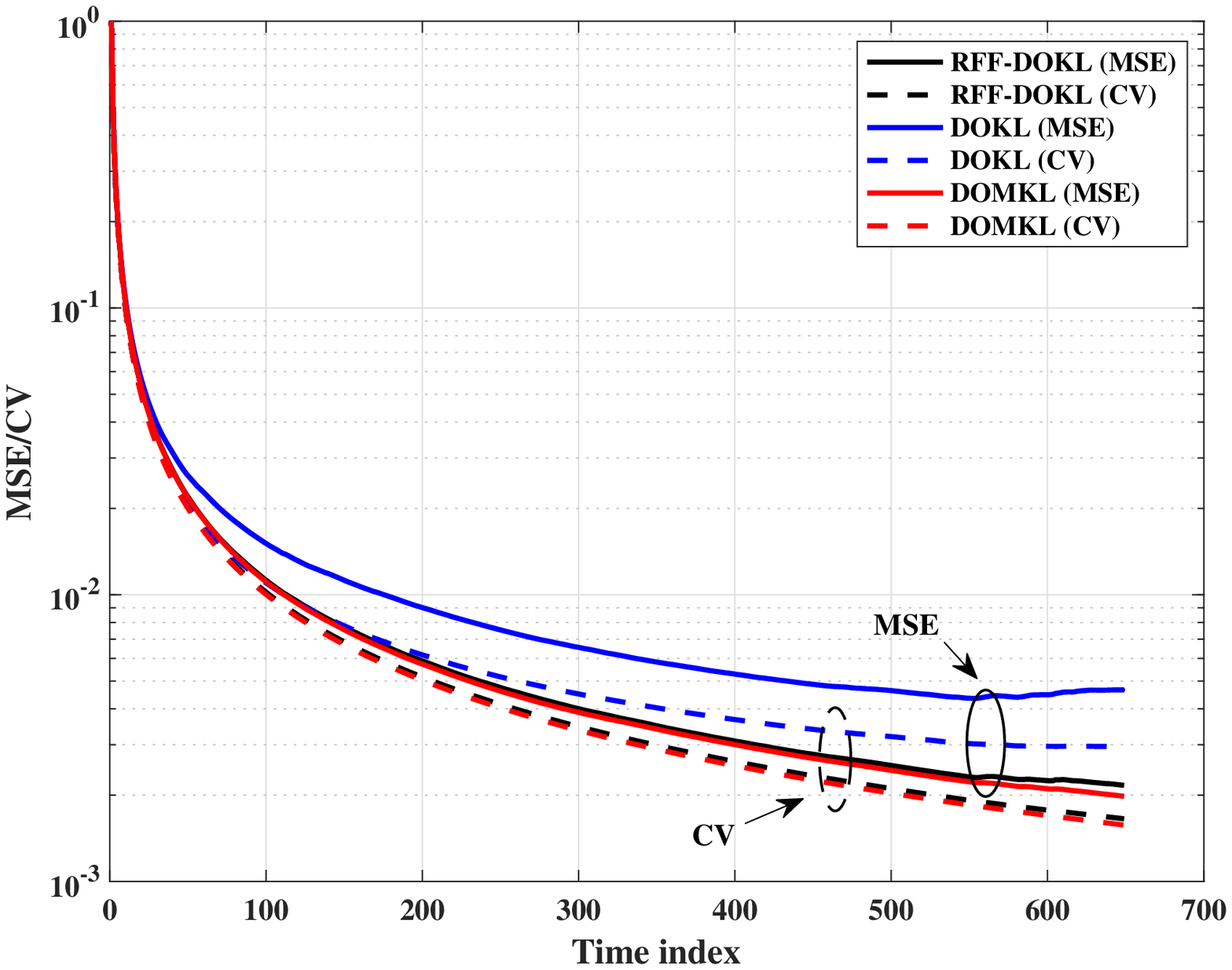}
}
\caption{Comparisons of MSE and CV performances of various methods in online regression ({\bf (a)} - {\bf (c)}) and time-series prediction ({\bf (d)} - {\bf (f)}) tasks. Here, a communication graph with $K=10$ learners is randomly constructed with a connection probability $\alpha_c = 0.25$.}
\label{fig:MSEperformance}
\end{figure*}

\subsubsection{Time-series prediction tasks}

For time-series prediction tasks, we consider the popular Autoregressive (AR) model \cite{mills1991time}. The AR($s$) model predicts the future value $y_{t}$ with the assumption of the linear dependency on its past $s$ values, which is formally defined as 
\begin{equation}
    y_{t} = c+\sum_{i=1}^{s}\gamma_{i}y_{t-i}+n_{t},
\end{equation} where $c$ is a constant, $\gamma_{i}$ denotes the weight associated with $y_{t-i}$, and $n_{t}$ denotes a Gaussian noise. Based on this, the RF-based kernelized AR($s$) model, which can explore a nonlinear dependency, has been introduced in \cite{hong2020active}, where it is formulated as
\begin{align}
y_{t}&=c+f(y_{t-1},y_{t-2},\dots,y_{t-s})+n_{t}\nonumber\\
&=c+f(\xv_{t})+\epsilon_{t},\label{eq:ARp}
\end{align} where $\xv_{t}\triangleq \left[y_{t-1},y_{t-2},\dots,y_{t-p}\right]^{\trasp}$ and $f(\xv_{t}) \in \bar{\Hc}$ (i.e., RKHS). In our experiments, the parameter $s$ is fixed as 5. Then, the above model can be directly plugged into the proposed and benchmark algorithms to solve time-series prediction tasks. These algorithms are tested with the following univariate time-series datasets from UCI Machine Learning Repository, which are also summarized in Table II:
\begin{itemize}
    \item {\bf Air quality} \cite{SDeVito2008} : This dataset includes 7322 time-series air quality data, where the features contain hourly response from an array of 5 metal oxide chemical sensors embedded in an Air Quality multi-sensor device deployed on the field in a city of Italy. The goal is to predict the concentration of polluting chemicals in the air.
    \item {\bf Traffic \cite{Timeseries}:} This dataset contains 6500 time-series traffic data obtained from Minneapolis Department of Transportation in US. Data is collected from hourly interstate 94 Westbound traffic volume for MN DoT ATR station 301.
    \item {\bf Temperature \cite{Timeseries}:} This dataset includes 5500 time-series temperature data obtained from Minneapolis Department of Transportation in US. Data is collected from hourly interstate 94 Westbound temperature for MN DoT ATR station 301, roughly midway between Minneapolis and St Paul, MN.
\end{itemize} { Unlike online regression tasks, the partition $\{\Dc_1,...,\Dc_K\}$ should keep the characteristic of a time-series data. This can be satisfied by building the partition in the following way.  Letting $\{(\xv_{t}, y_t): t\in [KT]\}$ be the time-series data in a dataset,  it is partitioned as $\Dc_{k}=\{(\xv_{k,t} = \xv_{K(t-1)+k}, y_{k,t}=  y_{K(t-1)+k}) , t \in [T]\}$ for $k \in \Vc$, where  $T=\left\lfloor \mbox{the total number of data}/K\right\rfloor$. Then, at time $t$, each learner $k$ receives the data $(\xv_{k,t}, y_{k,t})$ from $\Dc_k$.}

\begin{figure}
\centerline{\includegraphics[width=9.5cm]{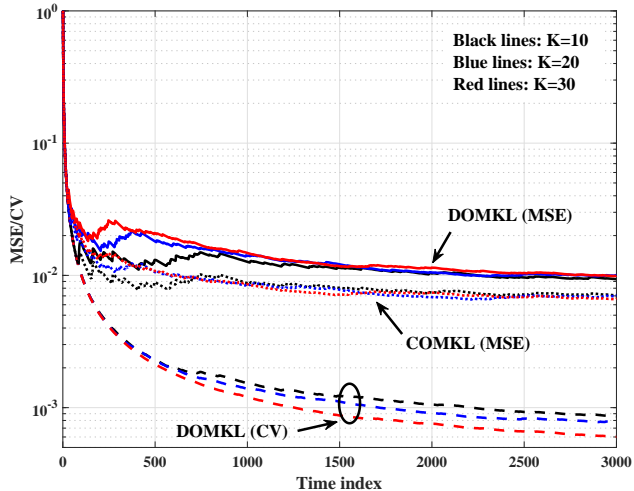}}
\caption{The MSE and CV performances of DOMKL and COMKL as a function of network size $K$ on online regression task with Twitter dataset, where a random communication graph is constructed with the connection probability $0.5$.} 
\label{fig:Networksize}
\end{figure}

%
%
{
\subsection{Performance Evaluations}\label{subsec:discuss}

We demonstrate the advantage of using multiple kernels with the comparison between DOMKL and the single kernel-based methods (e.g., DOKL and RFF-DOKL). The proposed DOMKL is first compared with DOKL with the best kernel, in which the best kernel was identified via an exhaustive search from the 17 Gaussian kernels in (\ref{eq:GK}), assuming that all incoming data is given in advance. As an example, it turns out that the Gaussian kernel with $\sigma^2=10^{-2}$ is the best kernel for the Wave energy data. We also confirmed that in this dataset, DOMKL yields the same MSE and CV performances of DOKL with $\sigma^2=10^{-2}$. The same results are also attained for the other datasets in Section~\ref{subsec:dataset}. These experiments reveal that as in OMKL \cite{hong2020active}, the proposed DOMKL can quickly find the best kernel in hindsight as the part of a learning process. In other words, our extension approach from DOKL to DOMKL is quite reasonable. We emphasize that in practice, it might be a chance to select an unsuitable single kernel, thus leading to a severe performance loss. This can be demonstrated via Fig.~\ref{fig:MSEperformance}, where in both DOKL and RFF-DOKL, the parameter of a Gaussian kernel is determined as either a higher bandwidth $\sigma^2=10^{-3}$ or a lower bandwidth $\sigma^2=10^3$. In Fig.~\ref{fig:MSEperformance}, the averaged performances of the above two cases are evaluated, namely, during the experiments, 250 trials with $\sigma^2 = 10^{-3}$ and the other 250 trials with $\sigma^2 = 10^3$ are performed. Hence, the performance gaps from DOMKL can be regarded as the performance loss in average caused by choosing an unsuitable single kernel. From Fig.~\ref{fig:MSEperformance}, we observe that the proposed DOMKL shows the best MSE and CV performances in all datasets, by enjoying the great advantage of using multiple kernels. Especially in the conductivity, weather, and traffic data, the performance gains obtained from using multiple kernels are tremendous.  This suggests the practicality of the proposed DOMKL.

In comparison between DOKL and RFF-DOKL, we can identify the difference of online ADMM and OGD-based diffusion strategy in the context of a distributed optimization. From the MSE performances in Fig.~\ref{fig:MSEperformance}, we observe that no algorithm is overwhelmingly better, i.e., DOKL can outperform RFF-DOKL and vice versa according to datasets. In contrast, DOKL shows better CV performances than RFF-DOKL in all datasets, implying that online ADMM seems to be more adequate for the network-wise consensus. Also, DOKL can be further elaborated by adjusting the hyper-parameters $\rho$ and $\eta_g$, whereas RFF-DOKL cannot. Nevertheless, it would be an interesting future work to extend RFF-DOKL in a multiple kernel setting (see Remark 4), since RFF-DOKL shows attractive performances in some datasets. 

Finally, we investigate the robustness of the proposed DOMKL in terms of the network scalability. From Fig.~\ref{fig:Networksize}, we observe that DOMKL yields the almost same MSE and CV performances regardless of the network size. Herein, a random communication graph is considered with a connection probability $\alpha_c = 0.5$. These results reveal that DOMKL performs uniformly well for various network sizes, thus being suitable for online learning tasks arising from massive IoT systems. We point out that the CV performances in Fig.~\ref{fig:Networksize} become better as the network size $K$ grows. This is because by construction, the number of neighboring learners can increase in average (e.g., $\alpha_c K$) as $K$ grows. Namely, as the network size increases, the updated information at each learner is propagated over a network more quickly, thus leading to a better consensus. Consider another type of a random communication graph with a fixed $\alpha_c K$ regardless of $K$ (e.g., $\alpha_c = \frac{c}{k}$ for some constant $c$). Differently from the previous case, the number of neighboring learners in average is fixed regardless of the network size $K$. As expected, one can observe the opposite trends of CV performances, compared with them in Fig.~\ref{fig:MSEperformance}. More importantly, in the both cases, we verified that CV values themselves are sufficiently small and also, the gaps between the best and worst CV values are almost negligible. Fig.~\ref{fig:MSEperformance} shows that DOMKL provides the comparable MSE performances with COMKL, in which the latter can be regarded as the lower bound. Namely, we can say that DOMKL almost achieves the optimal distributed online learning algorithm based on multiple kernels. Due to its attractive performance and the robustness on various network structures, thus,  the proposed DOMKL would be a promising candidate for fully decentralized online learning tasks. 
}

%

%
%
\section{Conclusion}\label{sec:con}

In this paper, we proposed a novel distributed online learning framework with multiple kernels (dubbed DOMKL). This method was devised by appropriately combining online alternating direction method of multipliers (online ADMM) and a distributed Hedge algorithm. The key advantages of the proposed DOMKL are the {\em scalability} in terms of the number of incoming data and {\em privacy-preserving}. As a theoretical contribution, we proved that the proposed DOMKL achieves the optimal sublinear regret, implying that it can yield the same asymptotic performance with the centralized OMKL, without centralizing the local data. Via experiments with real datasets, we demonstrated the effectiveness of the proposed DOMKL on various online regression and time-series prediction tasks. The advantages of scalability, privacy-preserving, and attractive performance suggest the practicability of DOMKL in real-world distributed online learning tasks. An interesting future work is to extend the proposed DOMKL into a more general communication network such as wireless and directed communication networks. Regarding wireless communication networks, the amount of exchanged information should be taken into account because of the limited capacity of a wireless channel. Active learning would be useful as it can only activate learners with an informative incoming data. This definitely reduces the communication overhead as well as the labeling cost. Another promising solution is to develop a quantized optimization technique suitable for our learning framework. As a result, exchanging quantized messages can reduce the communication overhead significantly.

\appendices

\section{Proof of Lemma~\ref{lem1-OADMM}}\label{app:lem1}
For ease of exposition, we will drop the subscript $p$ as the proof holds for any kernel $\kappa_p$. Accordingly, $\hat{\thetav}_{[t,p]}$ (resp., $\hat{\thetav}_{[k,t,p]}$) is replaced by $\hat{\thetav}_{t}$ (resp., $\hat{\thetav}_{[k,t]}$). This proof is based on online ADMM proposed in \cite{wang2013online}. To formulate the standard form of online ADMM, we let
\begin{equation*}
    \thetav_{[1:K]}\eqdef[\thetav_{1}^{\trasp},\thetav_{2}^{\trasp},\cdots,\thetav_{K}^{\trasp}]^{\trasp} \in \RR^{2MK\times 1},
\end{equation*}with $\thetav_{k} \in \RR^{2M\times 1},\; \forall k \in [K]$. Also, we introduce an auxiliary vector $\gammav_{\{k,\ell\}} \in \RR^{2M\times 1}$ denotes an auxiliary vector associated with an edge $\{k,\ell\}\in\Ec$. Then, the optimization problem in \eqref{eq:opt1} can be rewritten as
\begin{align}
      \hat{\thetav}_{t+1} &= \argmin_{\thetav_{[1:K]}}  \sum_{k=1}^{K} \Lc(\thetav_k^{\trasp}\zv_p(\xv_{k,t}), y_{k,t}) + \frac{\eta_l}{2}\|\thetav_{[1:K]} - \hat{\thetav}_{t}\|^2\nonumber\\
    &\mbox{ subject to } \Am\thetav_{[1:K]} + \Bm\gammav = \zerov, \label{eq:opt10}
\end{align} where $\Am \in \RR^{4M|\Ec|\times 2MK}$, $\Bm\in\RR^{4M|\Ec|\times 2M|\Ec|}$, and $\gammav=[\gammav_{\{k,\ell\}}: \{k,\ell\} \in \Ec] \in \RR^{2M|\Ec|\times 1}$ are specified below. {We will explain how to determine the $\Am$, $\Bm$, and $\gammav$ from a given network $\Gc=(\Vc,\Ec)$. Consider the simple graph $\Gc(\{1,2,3\},\{\{1,2\},\{1,3\}\})$. Given the set $\Ec=\{\{1,2\},\{1,3\}\}$, the auxiliary vector $\gammav$ is defined as 
\begin{equation*}
    \gammav = [\gammav_{\{1,2\}}^{\trasp},\gammav_{\{1,3\}}^{\trasp}]^{\trasp}.
\end{equation*} 
From the connectivity of the nodes, the matrices of $\Am$ and $\Bm$ are determined as
\begin{align}
    \Am &=\left[
    \begin{matrix} \Id_{2M} & \zerov_{2M} & \zerov_{2M} \\
    \Id_{2M} & \zerov_{2M} & \zerov_{2M}\\
    \zerov_{2M} &\Id_{2M} & \zerov_{2M}\\
    \zerov_{2M}&\zerov_{2M}&\Id_{2M}
    \end{matrix}\right] \mbox{ and }\Bm =- \left[\begin{matrix}
    \Id_{2M} & \zerov_{2M} \\
    \zerov_{2M} & \Id_{2M} \\
     \Id_{2M} & \zerov_{2M} \\
    \zerov_{2M} &  \Id_{2M}
    \end{matrix}
    \right],\label{eq:matrix}
\end{align} where $\Id_{2M}$ and $\zerov_{2M}$ represent the $2M\times 2M$ identify and all-zero matrices, respectively. Then, it is clearly verified that the consensus constraint in \eqref{eq:opt10} is equivalent to that in (\ref{eq:const_o2}) because
\begin{equation}\label{eq:example_A}
    \Am\thetav_{[1:3]} + \Bm\gammav = \left[\begin{matrix} \thetav_1 - \gammav_{\{1,2\}}\\
    \thetav_1 - \gammav_{\{1,3\}}\\
    \thetav_2 - \gammav_{\{1,2\}}\\
    \thetav_3 - \gammav_{\{1,3\}}
    \end{matrix}
    \right] = \zerov.
\end{equation} As shown in \eqref{eq:matrix}, $\Am$ and $\Bm$ are represented by simple block matrices consisting of $2M\times 2M$ identify and zero matrices. For ease of exposition, we let $(i,j)$ denote the indices of block-row of $\Am$ with $\{i,j\}=\{j,i\} \in \Ec$. In \eqref{eq:matrix}, we have the four block-rows whose are indexed by $(1,2)$, $(1,3)$, $(2,1)$, and $(3,1)$ in that order. Also, the block-columns of $\Am$ are indexed by numbers as usual. The matrix $\Am$ is constructed by determining the locations of $\Id_{2M}$ (i.e., non-zero matrices), for which $((i,j),k)$-th element (e.g., a block matrix) is equal to $\Id_{2M}$ only if $k = i$. The matrix $\Bm$ can be easily constructed using the circular shift of the first block-row $\Em$, where
\begin{equation}
    \Em\eqdef -\left[\begin{matrix} \Id_{2M} & \zerov_{2M} & \cdots & \zerov_{2M}\end{matrix}\right] \in \RR^{2M\times |\Ec|}.
\end{equation} In detail, the $i$-th block-row is obtained by the $i-1$ times circular shift of $\Em$ to the right.} 

The augmented Lagrangian of (\ref{eq:opt10}) is obtained as
\begin{align}
    &L_{\rho}^t(\thetav_{[1:K]},\gammav, \lambdav)\nonumber\\
    &\;\;\;\;\;\;\;=\sum_{k=1}^{K} \Lc(\thetav_k^{\trasp}\zv_p(\xv_{k,t}), y_{k,t}) + \lambdav^{\transp}(\Am\thetav_{[1:K]}+\Bm\gammav) \nonumber\\
    &\;\;\;\;\;\;\;+ \frac{\eta_l}{2}\|\thetav_{[1:K]}-\hat{\thetav}_{t}\|^2 + \frac{\rho}{2}\|\Am\thetav_{[1:K]}+\Bm\gammav\|^2.
\end{align} Online ADMM, then, consists of the three steps \cite{wang2013online}:
\begin{align}
    \hat{\thetav}_{t+1}&=\argmin_{\thetav_{[1:K]}} \sum_{k=1}^{K} \Lc(\thetav_k^{\trasp}\zv(\xv_{k,t}), y_{k,t}) + \hat{\lambdav}_t^{\trasp} (\Am\thetav_{[1:K]}+\Bm\hat{\gammav}_t) \nonumber\\
    &+ \frac{\rho}{2}\|\Am\thetav_{[1:K]}+\Bm\hat{\gammav}_t\|^2 + \frac{\eta_l}{2}\|\thetav_{[1:K]}-\hat{\thetav}_{t}\|^2,\label{eq:theta_j}\\
    \hat{\gammav}_{t+1} 
    &=\argmin_{\gammav} \hat{\lambdav}_t^{\transp}(\Am\hat{\thetav}_{t+1}+\Bm\gammav) +\frac{\rho}{2}\|\Am\hat{\thetav}_{t+1}+\Bm\gammav\|^2,\label{eq:zv}\\
    \hat{\lambdav}_{t+1}&=\hat{\lambdav}_t + \rho(\Am\hat{\thetav}_{t+1}+\Bm\hat{\gammav}_{t+1}).\label{eq:lambdav}
\end{align} As shown in \eqref{eq:example_A}, we have that
\begin{equation}
    \Am\thetav_{[1:K]} + \Bm\hat{\gammav}_t =
    \left[
    \begin{matrix}
   \cv_1\\
   \cv_2\\
   \vdots\\
   \cv_{K}
    \end{matrix}
    \right] \mbox{ with }  \cv_{k} = \left[ \begin{matrix} 
    \thetav_k - \hat{\gammav}_{\{k,k_1\},t}\\
    \vdots\\
    \thetav_k - \hat{\gammav}_{\{k,k_{|\Nc_k|}\},t}\\
    \end{matrix}
    \right],
    \label{eq:const-24}
\end{equation}
where $\{k_\ell$, $\ell \in \Nc_k\}$ denote the indices of the neighbors of the learner $k$. Also, $\hat{\lambdav}_{(k,\ell),t}$ denotes the components of $\hat{\lambdav}_{t}$ corresponding to the element $\thetav_{k} - \hat{\gammav}_{\{k,\ell\},t}$ in the $\Am\thetav_{[1:K]}+\Bm\hat{\gammav}_t$. Similarly, $\hat{\lambdav}_{(\ell,k),t}$ corresponds to $\thetav_{\ell} - \hat{\gammav}_{\{k,\ell\},t}$. Note $(k,\ell)$ denotes the ordered pair while $\{k,\ell\}$ denotes the unorderd pair. 
From (\ref{eq:const-24}) and using the above notations, the optimization problem in \eqref{eq:theta_j} can be decomposed into $K$ sub-optimization problems, in which the sub-optimization problem with respect to $\thetav_k$ is formulated as
\begin{align}
    &\hat{\thetav}_{[k,t+1]} =\argmin_{\thetav_k} \Lc(\thetav_k^{\trasp}\zv(\xv_{k,t}), y_{k,t}) + \hat{\lambdav}_{[k,t]}^{\trasp}\thetav_k \nonumber\\
    &\;\;\;\;\;\;\;\;\;\;+\frac{\rho}{2}\sum_{\ell\in\Nc_k}\|\thetav_k - \hat{\gammav}_{\{k,\ell\},t}\|^2 + \frac{\eta_l}{2}\|\thetav_k - \hat{\thetav}_{[k,t]}\|^2,\label{eq:thetav}
\end{align} where $\hat{\lambdav}_{[k,t]} = \sum_{\ell \in \Nc_k} \hat{\lambdav}_{(k,\ell),t}$. Likewise, the optimization problem in (\ref{eq:zv}) can be decomposed such as
\begin{align}
    &\hat{\gammav}_{\{k,\ell\},t+1} \nonumber\\
    &= \argmin_{\gammav_{\{k,\ell\}}} \hat{\lambdav}_{[k,t]}^{\trasp}(\hat{\thetav}_{[k,t+1]}-\gammav_{\{k,\ell\}})+ \hat{\lambdav}_{[\ell,t]}^{\trasp}(\hat{\thetav}_{[\ell,t+1]}-\gammav_{\{k,\ell\}})\nonumber\\
    &\;\;\;\;\; +\frac{\rho}{2}(\|\hat{\thetav}_{[k,t+1]}-\gammav_{\{k,\ell\}}\|^2+\|\hat{\thetav}_{[\ell,t+1]}-\gammav_{\{k,\ell\}}\|^2),\label{eq:54}
\end{align}for any $\{k,\ell\}\in \Ec$. The optimal solution of \eqref{eq:54} can be easily obtained as
\begin{align}
&\hat{\gammav}_{\{k,\ell\},t+1} =\frac{1}{2}(\hat{\thetav}_{[k,t+1]} + \hat{\thetav}_{[\ell,t+1]}) + \frac{1}{2\rho}(\hat{\lambdav}_{(k,\ell),t}+\hat{\lambdav}_{(\ell,k),t}).\label{eq:u_z}
\end{align} Also, we obtain that
\begin{align}
    &\hat{\lambdav}_{(\ell,k),t+1} = \hat{\lambdav}_{(\ell,k),t} + \rho(\hat{\thetav}_{[\ell,t+1]}- \hat{\gammav}_{\{k,\ell\},t+1})\label{eq:o_lambda}\\
    &\;\;\;\;\;\;\;\; = \frac{\rho}{2}(\hat{\thetav}_{[\ell,t+1]}-\hat{\thetav}_{[k,t+1]}) + \frac{1}{2}(\hat{\lambdav}_{(\ell,k),t} - \hat{\lambdav}_{(k,\ell),t}). \label{eq:u_lambda}
\end{align} From (\ref{eq:u_lambda}), the following equality holds:
\begin{equation}
    \hat{\lambdav}_{(\ell,k),t} + \hat{\lambdav}_{(k,\ell),t} = \zerov. \label{eq:prop}
\end{equation} Also, from (\ref{eq:u_z}) and (\ref{eq:prop}), we have:
\begin{align}
\hat{\gammav}_{\{k,\ell\},t+1} &=\frac{1}{2}(\hat{\thetav}_{[k,t+1]} + \hat{\thetav}_{[\ell,t+1]}).\label{eq:zv_2}
\end{align} By combining (\ref{eq:zv_2}) and (\ref{eq:thetav}), we obtain the (\ref{eq:theta-update}). From (\ref{eq:o_lambda}) and (\ref{eq:zv_2}), we have:
\begin{equation}
     \hat{\lambdav}_{(k,\ell),t+1} = \hat{\lambdav}_{(k,\ell),t} + \frac{\rho}{2} (\hat{\thetav}_{[k,t+1]} - \hat{\thetav}_{[\ell,t+1]}).
\end{equation} Thus, we can get:
\begin{align*}
    \hat{\lambdav}_{[k,t+1]}&=\sum_{\ell \in\Nc_k}\hat{\lambdav}_{(k,\ell),t+1}\\
    &=\hat{\lambdav}_{[k,t]} + \frac{\rho}{2}\sum_{\ell \in\Nc_k}(\hat{\thetav}_{[k,t+1]} - \hat{\thetav}_{[\ell,t+1]}).
\end{align*}
This completes the proof.

\section{Proof of Lemma~\ref{lem2}}\label{app:lem2}

From (\ref{eq:theta-update}) and (\ref{eq:lambda-update}), the gradient of the loss function can be represented as
\begin{align}
    &\nabla\Lc_{[k,t,p]}(\hat{\thetav}_{[k,t+1,p]}) \nonumber\\
    &= - \Big(\hat{\lambdav}_{[k,t+1,p]} + \frac{\rho}{2}\sum_{\ell \in \Nc_k}(\hat{\thetav}_{[k,t+1,p]}+\hat{\thetav}_{[\ell,t+1,p]})\nonumber\\
    &\;\;\;\; - \frac{\rho}{2}\sum_{\ell \in\Nc_k}(\hat{\thetav}_{[k,t,p]} + \hat{\thetav}_{[\ell,t,p]})+\eta_l(\hat{\thetav}_{[k,t+1,p]} - \hat{\thetav}_{[k,t,p]})\Big)\nonumber\\
    &= - \left(\hat{\lambdav}_{[k,t+1,p]} + \rho(\hat{\gammav}_{[k,t+1,p]} - \hat{\gammav}_{[k,t,p]})\right)\nonumber\\
    &\;\;\;\;- \eta_l \left(\hat{\thetav}_{[k,t+1,p]} - \hat{\thetav}_{[k,t,p]}\right),\label{eq:def_grad}
\end{align} where the last equality follows the definition of $\hat{\gammav}_{k,t}$. From the convexity of the loss function, we obtain:
\begin{align}
    &\Lc_{[k,t,p]} (\hat{\thetav}_{[k,t+1,p]}) - \Lc_{[k,t,p]}(\thetav_{[k,p]}^{\star})\nonumber\\
    &\leq \langle \nabla\Lc_{[k,t,p]}(\hat{\thetav}_{[k,t+1,p]}), \hat{\thetav}_{[k,t+1,p]} - \thetav_{[k,p]}^{\star}\rangle\nonumber\\
    &\stackrel{(a)}{=} - \langle\hat{\lambdav}_{[k,t+1,p]}, \hat{\thetav}_{[k,t+1,p]}-\thetav_{[k,p]}^{\star} \rangle\nonumber\\
    &\;\;\;\;\;+\rho\langle-\hat{\gammav}_{[k,t+1,p]}+\hat{\gammav}_{[k,t,p]}, \hat{\thetav}_{[k,t+1,p]} - \thetav_{[k,p]}^{\star}\rangle\nonumber\\
    &\;\;\;\;\; -\eta_l\langle\hat{\thetav}_{[k,t+1,p]} - \hat{\thetav}_{[k,t,p]}, \hat{\thetav}_{[k,t+1,p]}-\thetav_{[k,p]}^{\star} \rangle,\label{eq:proof1}
\end{align} where (a) is from (\ref{eq:def_grad}). Using the fact that 
\begin{align*}
    &\langle\vv_1 - \vv_2, \vv_3 + \vv_4\rangle\\
    &=\frac{1}{2}(\|\vv_4-\vv_2\|^2-\|\vv_4-\vv_1\|^2 +\|\vv_3+\vv_1\|^2-\|\vv_3+\vv_2\|^2),
\end{align*} we obtain the following inequality:
\begin{align}
    &\langle-\hat{\gammav}_{[k,t+1,p]}+\hat{\gammav}_{[k,t,p]}, \hat{\thetav}_{[k,t+1,p]} - \thetav_{[k,p]}^{\star}\rangle\nonumber \\
    &= \frac{1}{2}(\|-\thetav_{[k,p]}^{\star}+\hat{\gammav}_{[k,t,p]}\|^2 -\|-\thetav_{[k,p]}^{\star}+\hat{\gammav}_{[k,t+1,p]}\|^2\nonumber\\
    &+\|\hat{\thetav}_{[k,t+1,p]}-\hat{\gammav}_{[k,t+1,p]}\|^2-\|\hat{\thetav}_{[k,t+1,p]} - \hat{\gammav}_{[k,t,p]}\|^2).\label{eq:proof2}
\end{align} The following inequality is also satisfied:
\begin{align}
    &- \langle\hat{\lambdav}_{[k,t+1,p]}, \hat{\thetav}_{[k,t+1,p]}-\thetav_{[k,p]}^{\star} \rangle+\frac{\rho}{2} \|\hat{\thetav}_{j,t+1}-\hat{\gammav}_{[k,t+1,p]}\|^2 \nonumber\\
    &\stackrel{(a)}{\leq} - \langle\hat{\lambdav}_{[k,t+1,p]}, \hat{\thetav}_{[k,t+1,p]}-\thetav_{[k,p]}^{\star} \rangle\nonumber\\
    &\;\;\;\;\; -\langle\hat{\lambdav}_{[k,t+1,p]}, \thetav_{[k,p]}^{\star}- \hat{\gammav}_{[k,t+1,p]} \rangle +\frac{\rho}{2} \|\hat{\thetav}_{j,t+1}-\hat{\gammav}_{[k,t+1,p]}\|^2\nonumber\\
    &\stackrel{(b)}{=}  - \frac{1}{\rho}\langle\hat{\lambdav}_{[k,t+1,p]}, \hat{\lambdav}_{[k,t,p]} - \hat{\lambdav}_{[k,t+1,p]}\rangle \nonumber\\
    &\;\;\;\;\; + \frac{1}{2\rho}\|\hat{\lambdav}_{[k,t,p]} - \lambdav_{[k,t+1,p]}\|^2\nonumber\\
    &=\frac{1}{2\rho}(\|\hat{\lambdav}_{[k,t,p]}\|^2-\|\hat{\lambdav}_{[k,t+1,p]}\|^2),\label{eq:proof3}
\end{align}where (a) follows the fact that
\begin{equation}
\langle \hat{\lambdav}_{[k,t+1,p]}, \thetav_{[k,p]}^{\star} - \hat{\gammav}_{[k,t+1,p]} \rangle \leq 0,
\end{equation} which is shown in the proof of \cite[Lemma 2]{wang2013online} and (b) is due to the fact that
\begin{equation*}
    \hat{\thetav}_{[k,t+1,p]}-\hat{\gammav}_{[k,t+1,p]} =\frac{1}{\rho}( \hat{\lambdav}_{[k,t,p]} - \hat{\lambdav}_{[k,t+1,p]}).
\end{equation*} Finally, we have:
\begin{align}
    &-\langle\hat{\thetav}_{[k,t+1,p]} - \hat{\thetav}_{[k,t,p]}, \hat{\thetav}_{[k,t+1,p]}-\thetav_{[k,p]}^{\star} \rangle\nonumber\\
    &=\frac{1}{2}\|\hat{\thetav}_{[k,t,p]}-\thetav_{[k,p]}^{\star}\|^{2} -\frac{1}{2} \|\hat{\thetav}_{[k,t+1,p]}-\thetav_{[k,p]}^{\star}\|^{2} \nonumber\\
    &-\frac{1}{2}\|\hat{\thetav}_{[k,t+1,p]}-\hat{\thetav}_{[k,t,p]}\|^2.\label{eq:proof4}
\end{align}
By integrating (\ref{eq:proof1}), (\ref{eq:proof2}), (\ref{eq:proof3}), and (\ref{eq:proof4}), we complete the proof.

\section{Proof of Lemma~\ref{lem3}}\label{app:lem3}

We focus on any fixed $k \in \Vc$. Letting $\Jc\eqdef\{k\}\cup\{\Nc_k\}$, we define
\begin{equation*}
    \zeta \eqdef \sum_{t=1}^{T}\log\left[\sum_{p=1}^{P}\hat{q}_{[k,t,p]}\exp\left(-\frac{1}{\eta_g}\sum_{j \in \Jc}\Lc(\hat{f}_{[j,t,p]}(\xv_{j,t}),y_{j,t})\right)\right].
\end{equation*} The proof will be complete by deriving the upper and lower bounds of $\zeta$. We first obtain the upper-bound:
\begin{align}
    \zeta &\stackrel{(a)}{=}\sum_{t=1}^{T}\EE\left[\exp\left(-\frac{1}{\eta_g}\sum_{j\in\Jc}\Lc(\hat{f}_{[j,t,I]}(\xv_{j,t}),y_{j,t})\right)\right]\nonumber\\
    &\stackrel{(b)}{\leq} \sum_{j\in \Jc}\left(\sum_{t=1}^{T} -\frac{1}{\eta_g}\EE\left[\Lc(\hat{f}_{[j,t,I]}(\xv_{j,t}),y_{j,t})\right]\right) + \frac{|\Nc_{k}|T L_u^2}{8\eta_g^2},\label{lem3-1}
\end{align} where (a) is due to the fact that $I$ denotes a random variable with the probability mass function $(\hat{q}_{[k,t,1]},...,\hat{q}_{[k,t,P]})$ and (b) follows the Hoeffding inequality with the bounded random variable $\Lc(\hat{f}_{[j,t,I]}(\xv_{j,t}),y_{j,t})$ \cite{wainwright2019high}. Letting $\hat{W}_{[t,p]} \eqdef \hat{w}_{[k,t,p]}\cdot \prod_{\ell \in\Nc_k}\hat{w}_{[\ell,t.p]}$, we derive the lower-bound on $\zeta$:
\begin{align}
    \zeta &\stackrel{(a)}{=}\sum_{t=1}^{T}\log\left(\frac{\sum_{p=1}^{P}\hat{W}_{[t+1,p]}^{}}{\sum_{p=1}^{P}\hat{W}_{[t,p]}}\right)\nonumber\\
    &\stackrel{(b)}{=}\log\left(\sum_{p=1}^{P}\hat{W}_{[T+1,p]}^{}\right) - \log\left(\sum_{p=1}^{P}\hat{W}_{[1,p]}^{}\right)\nonumber\\
    &= \log\left(\sum_{p=1}^{P}\hat{W}_{[T+1,p]}^{}\right) - \log{P}\nonumber\\
    &\geq-\frac{1}{\eta_g}\sum_{j\in\Jc} \left(\min_{1\leq p\leq P}\sum_{t=1}^{T} \Lc(\hat{f}_{[j,t,p]}(\xv_{j,t}),y_{j,t})\right) - \log{P}, \label{lem3-2}
\end{align} where (a) follows the definition of $\hat{q}_{[k,t,p]}$ and $\hat{W}_{[t,p]}$ and (b) is from the telescoping sum. From (\ref{lem3-1}) and (\ref{lem3-2}), we have:
\begin{align*}
    -\frac{1}{\eta_g}\sum_{j\in\Jc} \left(\min_{1\leq p\leq P}\sum_{t=1}^{T} \Lc(\hat{f}_{[j,t,p]}^{}(\xv_{j,t}),y_{j,t})\right) - \log{P}\\
    \leq \sum_{j\in \Jc}\left(\sum_{t=1}^{T} -\frac{1}{\eta_g}\EE\left[\Lc(\hat{f}_{[j,t,I]}(\xv_{j,t}),y_{j,t})\right]\right) + \frac{|\Nc_k|T L_u^2}{8\eta_g^2}.
\end{align*} Rearranging the above inequality, we can get:
\begin{align*}
    &\sum_{t=1}^{T}\EE\left[\Lc(\hat{f}_{[k,t,I]}(\xv_{k,t}),y_{k,t})\right] - \min_{1\leq p\leq P}\sum_{t=1}^{T} \Lc(\hat{f}_{[k,t,p]}(\xv_{k,t}),y_{k,t})\\
    &\leq \sum_{\ell \in\Nc_k}\left[\min_{1\leq p\leq P} \sum_{t=1}^{T}\Lc(\hat{f}_{[\ell,t,p]}(\xv_{\ell,t}),y_{\ell,t})\right.\\
    &\left.-\sum_{t=1}^{T} \EE\left[\Lc\left(\hat{f}_{[\ell,t,I]}(\xv_{\ell,t}),y_{\ell,t}\right)\right] \right]  + \frac{ |\Nc_k|T L_u^2}{8\eta_g} + \eta_g\log{P}\\
    &\stackrel{(a)}{\leq} \frac{ |\Nc_k| T L_u^2}{8\eta_g} + \eta_g\log{P},
\end{align*} where (a) follows the fact that for any $\ell \in \Vc$,
\begin{equation*}
   \min_{ 1\leq p \leq P} \sum_{t=1}^T \Lc(\hat{f}_{[\ell,t,p]}(\xv_{\ell,t}),y_{\ell,t})\leq\sum_{t=1}^{T}\EE\left[\Lc(\hat{f}_{[\ell,t,I]}(\xv_{\ell,t}),y_{\ell,t})\right].
\end{equation*} Finally, setting $\eta_g = \Oc(\sqrt{T})$, the sublinear regret is achieved, which completes the proof.

\section*{Acknowledgment}
This work was supported by the National Research Foundation of Korea(NRF) grant funded by the Korea government(MSIT) (NRF-2020R1A2C1099836).


\ifCLASSOPTIONcaptionsoff
  \newpage
\fi


\bibliographystyle{IEEEtran}

\end{document}

%% file: macros.tex
\long\def\comment#1{}

\newfont{\bbb}{msbm10 scaled 700}

\newfont{\bb}{msbm10 scaled 1100}

\newcommand{\RR}{\mbox{\bb R}}

\newcommand{\EE}{\mbox{\bb E}}

\newcommand{\cv}{{\bf c}}

\newcommand{\vv}{{\bf v}}
\newcommand{\xv}{{\bf x}}

\newcommand{\zv}{{\bf z}}
\newcommand{\zerov}{{\bf 0}}

\newcommand{\Am}{{\bf A}}
\newcommand{\Bm}{{\bf B}}

\newcommand{\Em}{{\bf E}}

\newcommand{\Id}{{\bf I}}

\newcommand{\Dc}{{\cal D}}
\newcommand{\Ec}{{\cal E}}

\newcommand{\Gc}{{\cal G}}
\newcommand{\Hc}{{\cal H}}

\newcommand{\Jc}{{\cal J}}

\newcommand{\Lc}{{\cal L}}

\newcommand{\Nc}{{\cal N}}
\newcommand{\Oc}{{\cal O}}

\newcommand{\Vc}{{\cal V}}

\newcommand{\gammav}{\hbox{\boldmath$\gamma$}}

\newcommand{\lambdav}{\hbox{\boldmath$\lambda$}}

\newcommand{\muv}{\hbox{\boldmath$\mu$}}

\newcommand{\thetav}{\hbox{\boldmath$\theta$}}

\newcommand{\eqdef}{\stackrel{\Delta}{=}}

\newcommand{\trasp}{{\sf T}}
\newcommand{\transp}{{\sf T}}
